\documentclass[review,12pt]{elsarticle}

\usepackage{graphicx}
\usepackage{xcolor}
\usepackage{hyperref}
\usepackage{amsmath}
\usepackage{graphicx}
\usepackage{times}
\usepackage{latexsym}
\usepackage{datetime}
\usepackage{adjustbox}
\usepackage{booktabs}
\usepackage{natbib}
\setcitestyle{numbers}
\usepackage{subfiles}
\usepackage{tabularx}
\usepackage{amsfonts}
\usepackage{comment}
\usepackage{siunitx}
\usepackage{todonotes}
\usepackage{pdflscape}
\usepackage{todonotes}
\usepackage{multirow}
\usepackage{soul}
\usepackage{enumitem}
\usepackage{caption}
\usepackage{multirow}
\usepackage{booktabs}
\usepackage{sepfootnotes}
\usepackage{longtable}
\newcommand{\hlt}[1]{\textcolor{blue}{#1}}
\renewcommand{\hlt}[1]{\textnormal{#1}}
\usepackage{subfig}
\usepackage{float}

\usepackage{xcolor,colortbl}

\newcommand{\mc}[2]{\multicolumn{#1}{l}{#2}}
\definecolor{Gray}{gray}{0.85}
\definecolor{LightCyan}{rgb}{0.88,1,1}
\definecolor{corn}{rgb}{0.98, 0.93, 0.36}
\definecolor{naplesyellow}{rgb}{0.98, 0.85, 0.37}
\definecolor{lightyellow}{rgb}{1.0, 1.0, 0.88}
\definecolor{lightblue}{rgb}{0.31,0.34,0.34}
\definecolor{aliceblue}{rgb}{0.94, 0.97, 1.0}
\definecolor{blizzardblue}{rgb}{0.67, 0.9, 0.93}

\newcolumntype{a}{>{\columncolor{Gray}}c}
\newcolumntype{b}{>{\columncolor{white}}c}

\newcommand{\para}[1]{\paragraph{\textnormal{\textbf{#1}}}}
\newdate{date}{01}{07}{2023}

\renewcommand{\vec}[1]{\mathbf{#1}}

\newcommand{\uls}{\begin{itemize}[leftmargin=*]}
\newcommand{\ule}{\end{itemize}}
\newcommand{\ols}{\begin{enumerate}[leftmargin=*]}
\newcommand{\ole}{\end{enumerate}}

\newlength{\wdth}

\journal{Methods}

\begin{document}

\begin{frontmatter}



\title{\includegraphics[width=0.08\textwidth]{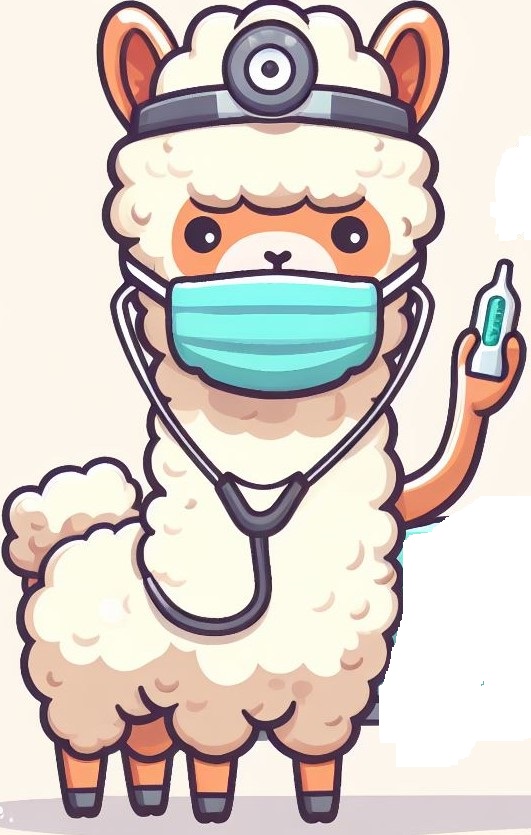} AlpaPICO: Extraction of PICO Frames from Clinical Trial Documents Using LLMs}
\author[smcs]{{Madhusudan Ghosh}$^\dagger$$^\star$}
\ead{madhusuda.iacs@gmail.com}
\author[smcs]{{Shrimon Mukherjee}$^\dagger$$^\star$}
\ead{iacsshrimon@gmail.com}
\author[IITP]{{Asmit Ganguly}$^{**}$}
\ead{asmitganguly.personal@gmail.com}
\author[smcs]{{Partha Basuchowdhuri}}
\ead{partha.basuchowdhuri@iacs.res.in}
\author[JU]{{Sudip Kumar Naskar}}
\ead{sudip.naskar@gmail.com}
\author[UoG]{{Debasis Ganguly}}
\ead{Debasis.Ganguly@glasgow.ac.uk}

\address{$\dagger$ The authors equally contributed to this paper.}
\address{$*$ Corresponding authors.}
\address{$**$ Work done during an internship at IACS.}

\address[smcs]{School of Mathematical and Computational Sciences, Indian Association for the Cultivation of Science}
\address[IITP]{Computer Science and Engineering, Indian Institute of Technology, Patna}
\address[UoG]{School of Computing Science,  University of Glasgow}
\address[JU]{Department of Computer Science and Engineering, Jadavpur University}

\begin{abstract}
In recent years, there has been a surge in the publication of clinical trial reports, making it challenging to conduct systematic reviews. Automatically extracting Population, Intervention, Comparator, and Outcome (PICO) from clinical trial studies can alleviate the traditionally time-consuming process of manually scrutinizing systematic reviews. Existing approaches of PICO frame extraction involves supervised approach that relies on the existence of manually annotated data points in the form of BIO label tagging. Recent approaches, such as In-Context Learning (ICL), which has been shown to be effective for a number of downstream NLP tasks, require the use of labeled examples. In this work, we adopt ICL strategy by employing the pretrained knowledge of Large Language Models (LLMs), gathered during the pretraining phase of an LLM, to automatically extract the PICO-related terminologies from clinical trial documents in unsupervised set up to bypass the availability of large number of annotated data instances. Additionally, to showcase the highest effectiveness of LLM in oracle scenario where large number of annotated samples are available, we adopt the instruction tuning strategy by employing Low Rank Adaptation (LORA) to conduct the training of gigantic model in low resource environment for the PICO frame extraction task.
More specifically, both of the proposed frameworks utilize AlpaCare as base LLM which employs both few-shot in-context learning and instruction tuning techniques to extract PICO-related terms from the clinical trial reports. We applied these approaches to the widely used coarse-grained datasets such as EBM-NLP, EBM-COMET and fine-grained datasets such as EBM-NLP$_{rev}$ and EBM-NLP$_{h}$. Our empirical results show that our proposed ICL-based framework produces comparable results on all the version of EBM-NLP datasets and the proposed instruction tuned version of our framework produces state-of-the-art results on all the different EBM-NLP datasets. Our project is available at \url{https://github.com/shrimonmuke0202/AlpaPICO.git}.
\end{abstract}

\end{frontmatter}


\section{Introduction}
\hlt{In the last few decades, the concept of Evidence-Based Medicine (EBM) has garnered significant interest within the healthcare community.} More specifically, EBM is a technique used by medical practitioners and healthcare professionals to guide their clinical decision-making regarding patient care by utilizing the highest quality and most up-to-date research evidence available~\cite{sackett1997evidence}.
Additionally, meta analysis is one of the necessary statistical technique in evidence synthesis literature to provide a sufficient number of necessary medical evidences by combining the results of different research studies to determine the necessary action~\cite{cook1997systematic}. Meta analysis is highly labor intensive and time-consuming process, due to the necessity for manually scrutinizing an extensive number of research articles and extracting pertinent information from them~\cite{jonnalagadda2015automating}. Recently, a general trend observed in the scientific literature of any discipline is that it grows at a rapid rate embracing new theories. The steep rise of scientific publications makes it difficult to manually conduct evidence synthesis. The process of systematically reviewing clinical data, including prescriptions and electronic health records, can be made simpler by automatically extracting relevant outcome terms. Previous research has not extensively explored the natural language processing (NLP)-based evidence synthesis literature due to the scarcity of annotated data~\cite{boudin2010positional,marshall-etal-2017-automating} to employ the relevant machine learning approaches for extracting the important components such as Participants/Populations (P), Interventions (I)/ Comparators (C)~\footnote[1]{It is to be noted that I and C are very often merged together into just I~\cite{nye-etal-2018-corpus,jin2018pico,kim2011automatic}.} and Outcomes (O)~\cite{huang2006evaluation}, also popularly known as PICO. \hlt{To alleviate this challenge, Nye et. al.~\cite{nye-etal-2018-corpus} developed an EBM-NLP dataset where the dataset uses an arbitrary selection process for outcome label annotations~\cite{abaho2019correcting}, and later on, revised datasets such as EBM-COMET and EBM-NLP-revised (EBM-NLP$_{rev}$) have been introduced in this literature}~\cite{abaho2022assessment}. \hlt{Additionally, to enhance the overall performance of the clinical trial task, the biomedical literature has witnessed a significant number of state-of-the-art (SOTA) pretrained models (PLMs) such as BNER~\cite{stubbs2015automated,uzuner2007evaluating} to conduct the biomedical named entity recognition task. However, such language models often struggle due to the scarcity of extensive annotated data instances. Moreover, the biomedical literature has seen a proliferation of publications aiming to leverage the pretrained knowledge of large language models (LLMs), gathered during the pretraining phase, such as PMC-Llama~\cite{wu2023pmcllama}, BioMedGPT-LM~\cite{luo2023biomedgpt}, etc., by fine-tuning them on domain specific tasks.}

Recently, AlpaCare~\cite{zhang2023alpacare} applied finetuning strategy on all the layers of gigantic Llama 2 chat model using a newly constructed medical instruction response dataset MedInstruct-52k. \hlt{However, this technique is highly resource intensive and time consuming due to the update process of larger number of parameters than that of in Llama 2 chat model.} \hlt{The main problem with this approach is that conducting domain specific training of these generative models is highly resource intensive and time consuming.} To bridge this gap, we apply a novel in-context learning (ICL) framework where an additional annotated context is provided to a LLM is shown to be effective for downstream PICO frame extraction task by bypassing the additional training process required in supervised setup. The context supplied is a set of annotated sentences, extracted from a relevant training corpus available for this downstream PICO frame extraction task. 
\begin{figure}[!t]%
    \centering
    % \vspace*{-1mm}
        {\includegraphics[width=.65\textwidth]{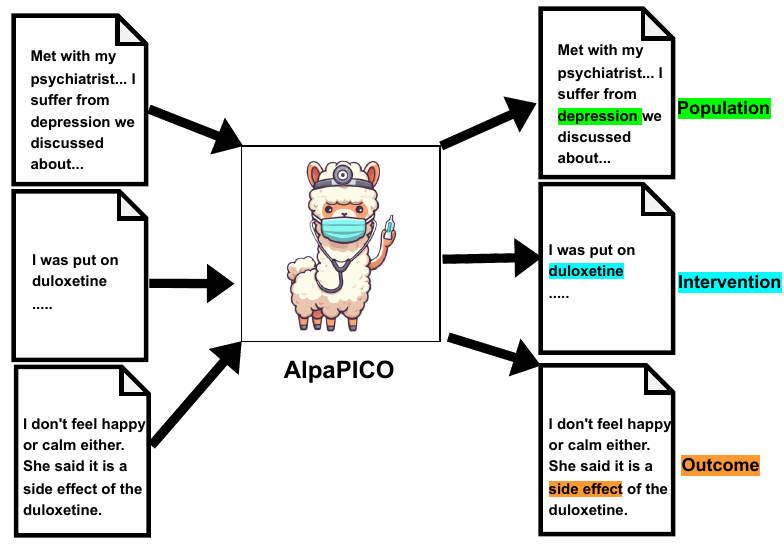}}
    \caption{\small Example of PICO frame extraction using our proposed framework AlpaPICO. Here we pass documents to our proposed framework for extracting PICO frames from the clinical trial documents.}
    \label{fig:intro}
\end{figure}
Additionally, in contrast to the existing supervised approaches~\cite{beltagy-etal-2019-scibert, yasunaga-etal-2022-linkbert}, by considering the PICO frame extraction as sequence classification task, we utilize parameter efficient finetuning strategy (PEFT) specifically by utilizing the low rank adaptation module (LoRa) to finetune AlpaCare on PICO frame generation task in low resource scenario. To the best of our knowledge, we are the first one to investigate the feasibility of applying both incontext learning (ICL) and instruction tuning strategies for PICO frame extraction task. The overall workflow of our framework has been shown in Figure~\ref{fig:intro}.

\subsection{Our Contributions}
To summarize, the following are our contributions in this paper.
\begin{enumerate}
\item To the best of our knowledge, we are first in exploring the potential of employing an ICL-based framework to perform the downstream PICO frame extraction task from biomedical literature by utilizing the pretrained knowledge of LLM, which effectively omits the entire training process of a supervised setup.

\item Our empirical results clearly demonstrate that our $k$-shot contexts in ICL-based framework significantly enhances the performance of $k$-shot ICL framework, in compare to the zero-shot scenario where the need for training is completely eliminated.

\item We also employ instruction tuning based approach to conduct the PICO frame extraction task on both the fine-grained and coarse-grained datasets available in EBM-NLP literature.

\end{enumerate}

\section{Related Work}
Our research focuses on examining the deep learning-based entity extraction approaches from the scientific literature. In this section, we breifly describe the state-of-the-art methodologies employed for named-entity recognition (NER) tasks in scientific domain.
\para{General NER Techniques}
NER is a very popular conventional subtask of information extraction in the NLP domain. Many techniques have been developed and are still being used in this field over the years. Some of the most important and recent ones are contextual embedding based model \cite{hoory2021learning},  BiLSTM-CRF \cite{mayhew2020robust}, CNN-based \cite{sung2021cnnbif}, Cross-BiLSTM-CNN and Att-BiLSTM \cite{li2020attention}, gated relation network by capturing global context~\cite{chen2019grn}. Xu et al. and Li et al. examined the performance of nested NER~\cite{xu2021supervised,li2020recursively}. Numerous studies are conducted on the joint entities and relations extraction~\cite{dai2019joint,zeng2020copymtl,nayak2020effective,xiao2020joint,sun2021progressive,li2021joint}, named entity normalization.~\cite{ji-etal-2021-neural}, and NER for low-resource languages~\cite{DBLP:journals/talip/DasGG17,mukherjee-etal-2023-mllab4cs}. 
Researchers have worked on important variants of NER task, such as, document level NER using a multitask learning approach~\cite{wang2021learning}, NER with a multi-level topic-aware attention mechanism~\cite{ma2023sequence} and information extraction using a multi-modal approach~\cite{toledo2019information}.  A popular application of NER is to extract scientific concept names from scientific literature~\cite{ghosh-etal-2022-astro,luan2018multi,jain2020scirex}. Some researchers have extracted AI methodology, and components from AI domain~\cite{ghosh2023extracting,ghosh2023enhancing}. Recently, NER has been applied in bio-NLP and biomedical domains~\cite{tong2021multi,xu2021supervised,li2020recursively,zhu2015aligning}.  Additionally, FLAIR~\citep{akbik-etal-2018-contextual}, a neural framework, helps significantly to finetune existing PLMs on downstream NER task to yield SOTA results.

\para{Evidence-based Medical Entity Extraction}
Extraction of PICO related terminologies from clinical trial text is an important area of research.
According to earlier research, the entity extraction task from bio-medical literature was performed at the sentence level when a large number of annotated data instances were unavailable~\cite{boudin2010combining,huang2011classification}. Recent inventions of pre-trained language models (PLMs) such as ELMo~\cite{peters-etal-2018-deep}, GPT~\cite{radford2018improving}, BERT~\cite{devlin-etal-2019-bert}, XLM~\cite{lample2019cross} and XLMNet~\cite{yang2019xlnet} etc. help significantly to mitigate the problem of limited annotated data samples. Such models also help to achieve state-of-the-art results on different NLP tasks including named entity recognition~\cite{sang2003introduction,rajpurkar2016squad}. Some recent studies \cite{yasunaga-etal-2022-linkbert,liu-etal-2021-sent2span-span,abaho2022assessment,beltagy-etal-2019-scibert} utilized PLMs for biological entity extraction task on the available datasets such as EBM-NLP~\cite{nye-etal-2018-corpus} corpus and EBP-NLP$_{rev}$~\cite{abaho2022assessment}.
The prior state-of-the-art models ~\cite{brockmeier2019improving,beltagy-etal-2019-scibert,zhang2020unlocking,ghosh2024blinktextsubscriptlstm}  on PICO entity recognition task performed poorly on the EBM-NLP corpus because it contains pharmaceutical intervention classes over non-pharmaceutical ones. Small-scale annotation resulted poorly in PICO span extraction task from clinical trial literature. Later researchers used distantly supervised datasets to overcome the problem of small annotated datasets~\cite{dhrangadhariya-muller-2022-distant,giannakopoulos-etal-2017-unsupervised}. Additionally, the PICO-related entity extraction task was performed by breaking down the available entity classes into different binary classes~\cite{mullenbach-etal-2018-explainable}.
\para{Large Language Models and In-context Learning} 
Recently LLMs~\cite{brown2020language,rae2021scaling,smith2022using,hoffmann2022training,chowdhery2022palm} have obtained significant improvement on a variety of NLP tasks~\cite{hegselmann2022tabllm,vilar2022prompting,perez2021true,pietrzak2021story,wei2021finetuned}. The use of LLMs for downstream tasks can be divided into two categories: firstly finetuning, and secondly in-context learning (ICL). In the finetuning strategy, a pretrained model is initialized, and additional epochs are executed on the downstream supervised data\cite{raffel2020exploring,gururangan2018annotation,roberts2020much,guu2020retrieval}. In contrast to that, ICL based strategy involves instructing LLMs to generate text based on few-shot demonstrations. Reformulating the first step of the downstream task involves incorporating prompts with demonstrations~\cite{radford2019language}. A systematic analysis of in-context learning framework was performed on various tasks by the GPT-3 model~\cite{brown2020language}. Chowdhery et. al.~\cite{chowdhery2022palm} performed analysis for the NMT task on PaLM. Researchers have shown that better prompts and demonstrations lead to a performance boost for in-context learning ~\cite{perez2021true,lu2021fantastically,rubin2021learning}. Recently, in-context learning based techniques have been used for NER task~\cite{wang2023gpt,zhang2023promptner,pakhale2023comprehensive,ashok2023promptner}.   
\para{Instruction tuning}
As a successful approach for customizing language models to handle diverse tasks, instruction tuning has garnered growing attention and engagement from the community. FLAN~\cite{chung2022scaling}, T0~\cite{sanh2021multitask}, and Tk-Instruct~\cite{wang2022super} transform extensive sets of pre-existing supervised learning datasets into an instruction-following format. Subsequently, they finetune encoder-decoder models, demonstrating robust zero-shot and few-shot performance across various NLP benchmarks. Researchers utilized crowd-sourced high-quality instructional data to finetune GPT-3, transforming it into InstructGPT and improving its capacity to comprehend user intentions and adhere to instructions~\cite{ouyang2022training}. Notably, recent progress in smaller models~\cite{alpaca,vicuna-2023,peng2023instruction} has demonstrated task-following capabilities, achieved through finetuning on instruction data generated by language models like ChatGPT or GPT-4. Nevertheless, smaller models frequently encounter difficulties in producing top-notch responses across various tasks~\cite{wang2023far}. A more detailed analysis of specific benchmarks exposes a significant disparity between these models and ChatGPT~\cite{gudibande2023false}. The study conducted by \cite{wang2023instructuie} investigates instruction-tuning for information extraction tasks. Nevertheless, their approach heavily depends on supervised datasets and demonstrates inferior performance compared to ChatGPT. Some works emphasized tuning models to excel at a specific type of task~\cite{zhou2023universalner}. The diversity in the instruction-tuning method is derived from task labels (e.g., relation types for relation extraction, entity types for NER), rather than instructions. By concentrating on task-level capabilities and employing NER as a case study, it is shown that a tuning recipe can be devised, not only closing down the performance gap but also surpassing the performance of ChatGPT. In this study~\cite{zhang2023alpacare}, the significance of task diversity in instruction tuning for the medical domain is demonstrated. Comprehensive experiments are conducted to evaluate free-form instruction in both medical and general domains. The results show that tuning AlpaCare with a diverse medical self-instruct dataset can simultaneously improve its medical capacity and generalization ability. Additionally, we introduce MedInstruct-52K, a diverse medical task dataset containing 52,000 instruction-response pairs, and MedInstruct-test, a set of novel medical tasks crafted by clinicians. These datasets aim to facilitate the development and evaluation of future domain-specific instruction-following models.

In this work, we are exploring an ICL-based framework to perform downstream PICO framework extraction tasks from clinical trial documents by using the knowledge of pretrained LLM, which omits the entire training process of a supervised setup. Additionally, we also perform an instruction tuning strategy to conduct the said task.

\section{Methodology}
In this section, we discuss the task definition followed by our proposed frameworks to conduct the downstream PICO frame extraction task.
\subsection{\hlt{Overview of Sequence Labeling task}}
\hlt{Given a clinical trial document ($D$), our goal is to identify the entity spans $(E_s)$ as well as the corresponding categories $(E_{PI/CO})$ of the identified entities from the sentences $D=(s_1,s_2,....,s_N)$ of that particular document.} In general, this task is considered as a traditional sequence labeling problem such that, given a sequence of words in any sentence $s_i = (w_0,w_1,....,w_N)$ from a sentence $s_i$, a supervised neural network model learns its parameters ($f_\theta$) to map an input sequence. Formally speaking, let $\vec{s}$ be a sentence of maximum length $N$ in a training set $\mathcal{T}$, i.e., $\vec{s} =\{w_0,\ldots,w_{N-1}\}$, where each $w_j$ is a token of that sentence. Let the set of mentions of entities occurring in $\vec{s}$ be $e(\vec{s}) = \{e_0,\ldots,e_{n-1}\} \subset \mathcal{E}$, where $\mathcal{E}$ contains the collection of PICO spans annotated in the dataset. Each mention, $e(\vec{s})$, is a subsequence of $\vec{x}$, which we denote by an indicator sequence of positions $\{\mathbb{I}(w_0), \ldots, \mathbb{I}(w_{N-1})\}$, where $\mathbb{I}(w_i) = 1$ if $w_i$ is a part of some entity $e \in e(\vec{x})$. A continuous span $\{j, \ldots, j+n-1\}$ (where $j \in \mathbb{Z}_M$) such that $\mathbb{I}(x_i)=1$, $\forall j \leq i \leq j+n-1$ denotes an entity comprised of $n$ tokens.  To distinguish the start of a span from its continuity and also its end, it is a common practice to denote the label of the first element of such an index set with $B$ (denoting Beginning of a span), the subsequent elements as $I$ (denoting that these are Inside a span) and the first index after the span ends.  Thus, each token sequence $\vec{s} \in \mathcal{T}$ of length $N$ is mapped to a label sequence of the same length, i.e., $\vec{s} = \{s_0,\ldots,s_{N-1}\} \mapsto \vec{y} = \{y_0,\ldots y_{N-1}\}$ where each $y_i \in \{B, I, O\}$. \hlt{Finally, given a set of examples (s,y) of such sequence pairs, the parameters $\theta$ of a sequence classification model are learned by optimizing
\begin{equation}
z = \underset{\theta}{\mathrm{argmin}}\, \sum_{(x, y)\in \mathcal{D}}\mathcal{L}(y,f(s,\theta)),
\label{eq:problem_definition_wloss}
\end{equation}
where $\mathcal{L}$ is a standard loss function, e.g., the cross-entropy.}

\begin{figure*}[!t]%
    \centering
    \hspace*{-10.5mm}
    {\includegraphics[width=0.95\textwidth]{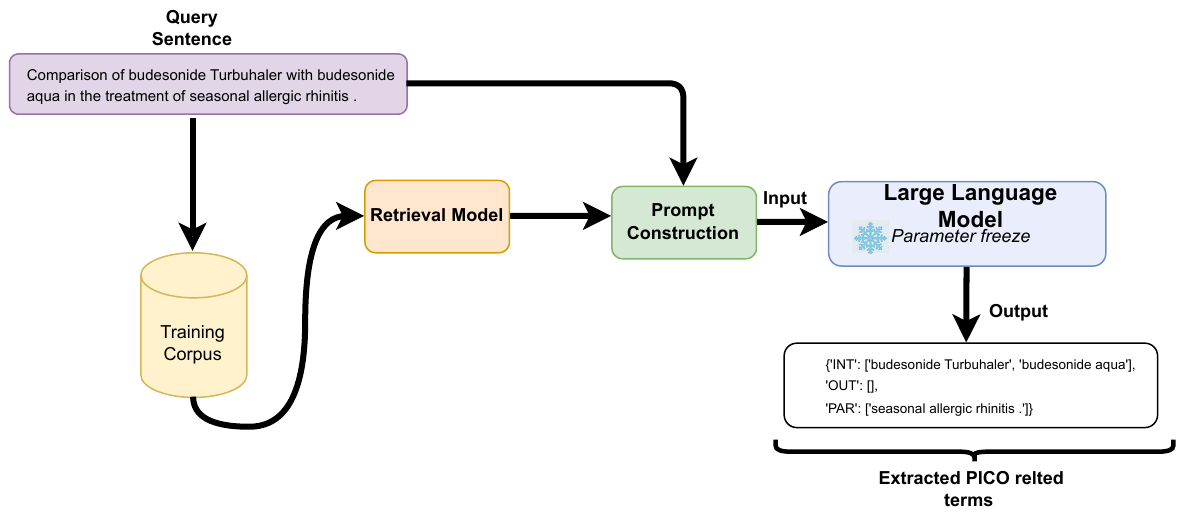}}
    % \captionsetup{width=0.7\textwidth}
     \vspace{-0.5em}
    \caption{\small Visuaization of our proposed ICL-based framework for PICO frame extraction}
    \label{fig:icl_architecture}
     % \vspace{-1.5em}
\end{figure*}
\subsection{Overview of In-Context Learning}
\label{ss:icl_framework}
Now we describe one of our proposed methodology to extract the entities from clinical trial text. In contrast to supervised learning, in-context learning (ICL) does not necessitate the training of a specified set of parameters $\theta$ on labeled instances. Instead, the posterior probabilities are now dependent on various factors, including a collection of $k$ labeled input examples, the decoder parameters of a pre-trained LLM. Our work investigates the feasibility of applying in-context learning (ICL) framework to extract the named entity recognition (NER) from the clinical literature to utilize the pretrained knowledge of LLM. The input $X = [T; D; I]$ incorporates the task description $T$, demonstrations $D$, and input sample $I$ while the generated output is a set of extracted entities of different categories $Y=\{E1_{type}:[e_1,...,e_n] ..., E2_{type}:[e_1,..,e_n]\}$. \hlt{Figure~\ref{fig:icl_architecture} presents an example from clinical trial dataset, where an ICL-based framework generates the biomedical entities of different types such as {``seasonal allergic rhinitis" as \emph{Intervention}, ``budesonide" as \emph{Participation} and \emph{Outcome}}, by utilizing the knowledge from the training instances. The overall ICL framework is depicted in Figure~\ref{fig:icl_architecture}.}
\begin{figure*}[ht]%
    \centering
    \hspace*{-10.5mm}
    {\includegraphics[width=0.95\textwidth]{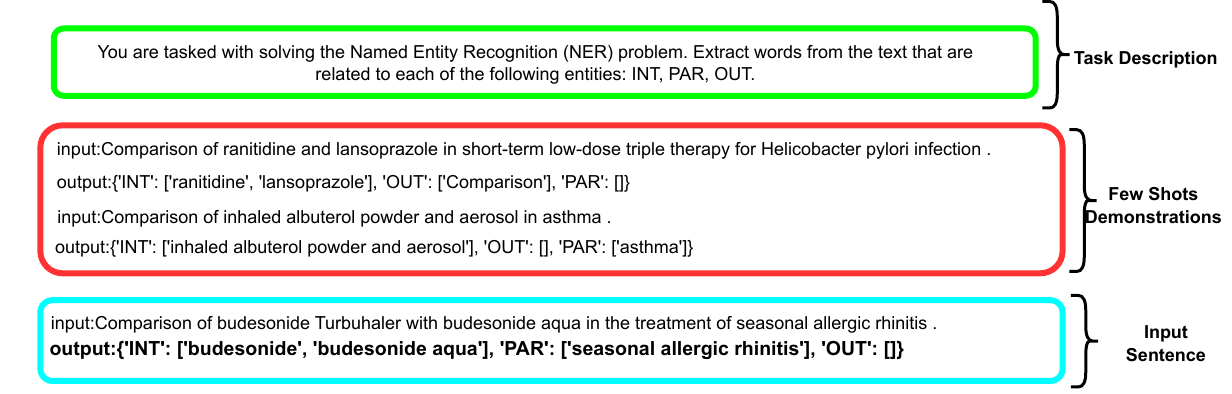}}
    % \captionsetup{width=0.7\textwidth}
     \vspace{-0.5em}
    \caption{\small Formats of input and output of our in-context learning based framework for name entity recognition task. The input is formed by the Task Description, Demonstrations, and the Input Sentence.}
    \label{fig:io}
     % \vspace{-1.5em}
\end{figure*}

\paragraph{\textbf{Task Description}} The task description provides an overview of entity recognition problem, with a specific focus on identifying `PICO' frames from clinical trial literature. This framework can be employed in the context of evidence-based medicine research, encompassing four key components: `Patient/Population', `Intervention', `Comparison', and `Outcome'. To perform `PICO' frame extraction task by utilizing the pretrained knowledge of LLM, it is important to guide LLM as discussed in the work of Min et al.~\citep{min-etal-2022-metaicl}. Additionally, Figure~\ref{fig:io} visually demonstrates the importance of the task description in helping the LLM to generate the three distinct types of entities: `Partition', `Intervention', and `Outcome'. \hlt{By following the outlined task description, the model acquires a comprehensive grasp of the intricacies associated with pinpointing and classifying the distinct entities encapsulated within the 'PICO' framework. This profound comprehension is pivotal for the model's ability to precisely decode and handle information pertinent to the specified task.}
\paragraph{\textbf{Demonstrations}}
Demonstrations play a vital role in ICL framework by conveying intra-class knowledge related to target entity types. This includes insights into entity semantics and contextual patterns, resulting in a comprehensive understanding of the subject matter. The essence of demonstrations is captured in Figure~\ref{fig:io}, where the demonstration instance within the illustrative set follows a specific template such as: ``input: \{text\} output: \{extractions\}." In this context, \{text\} demonstrates the semantically similar context relevant to the target sentence from where we need to identify the to the textual content being considered, while \{extractions\} represents the extracted entities present in the given text. This distinction facilitates a clear and detailed representation of the output format, enabling a deeper comprehension of the target entity types and their contextual representations.

\paragraph{\textbf{Extractions}}
The outcome of the extraction procedure yields an array of entities, with each distinct extraction articulated as "\textsc{Entity} is \textit{type}". For instance, as illustrated in Figure~\ref{fig:io}, the extraction of "treatment of seasonal allergic rhinitis" denotes "allergic rhinitis" as an identified entity classified under "Participation". This representation style, akin to natural language, facilitates the efficient leverage of the inherent text generation proficiencies of large language models by tapping into the extensive knowledge they have amassed during the pretraining phase.
\para{Architecture}
In line with the task formulation, we have chosen to employ LLaMA based models, namely, AlpaCare~\citep{zhang2023alpacareinstructiontuned}.
\hlt{AlpaCare is specifically developed for medical applications while the primary goal of AlpaCare is to enhance the model's proficiency in the medical domain while maintaining strong generalization capabilities across various tasks. The decoder of AlpaCare framework is tasked with processing diverse inputs, such as instructions, demonstrations, and textual data.}  These inputs are represented as demonstrations. Subsequently, the decoder functions to generate a comprehensive set of extractions, which take the form of a tokenized text sequence denoted as $Y=[y_1, \dots, y_n]$.

\hlt{The efficacy of an ICL-based framework primarily depends upon two crucial capabilities, namely, the ability to learn contextually and extract suitable information.} In this manner, we can perceive a \hlt{Large} Language Model (LLM) as a meta-function, that is, a function that employs extractors as its input (in the form of instructions and demonstrations) and produces the requisite entity extractor as its output. \hlt{To choose the appropriate text portion, we employ dense vector representations from BioBERT to calculate the cosine similarity.}

\para{Retrieval-based approach to obtain relevant context for Entity Recognition Task}
To obtain the set of text units for our downstream PICO frame extraction task, we first execute a keyword query formulated from the input sentences of training data on a dense index constructed from the document collection. \hlt{As the granularity of retrievable units, we work at the sentence level to match the input sentence's length.}
A dense index retriever, which we used in our experiments similar to \cite{wu-etal-2023-openicl}, outputs a list of top-$k$ candidate sentences retrieved from an index. \hlt{Our proposed ICL-based framework for the PICO frame extraction task is depicted in Figure~\ref{fig:icl_architecture}.}

\section{Downstream Task focused Instruction Tuning}
\label{ss:inst_tuning}
Additionally, to the best of our knowledge, we are the first to propose an instruction tuning approach along with an ICL-based framework which helps to improve the zero-shot and few-shot performance of LLMs such as Alpaca~\cite{alpaca} and Vicuna~\cite{vicuna-2023}.
\begin{figure*}[!t]%
    \centering
    \hspace*{-10.5mm}
    {\includegraphics[width=0.55\textwidth]{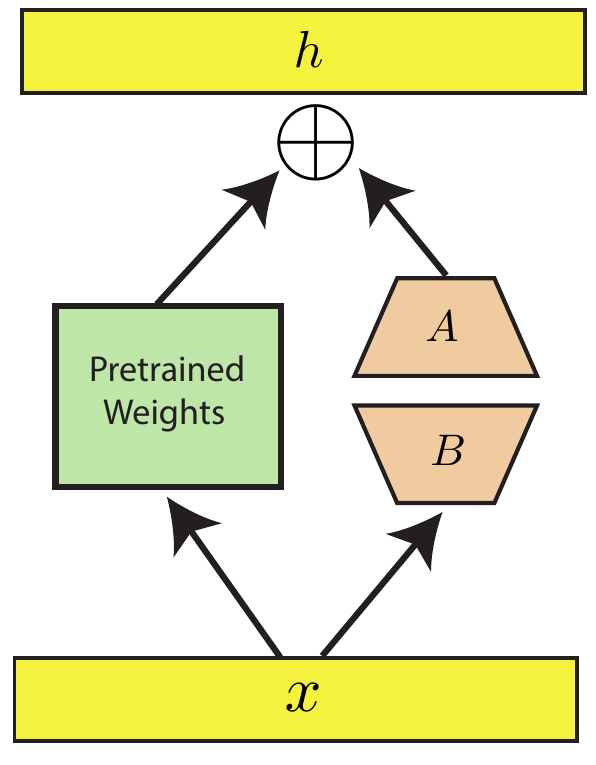}}
    % \captionsetup{width=0.7\textwidth}
     \vspace{-0.5em}
    \caption{\small LoRA block of our AlpaPICO framework.}
    \label{fig:lora}
     % \vspace{-1.5em}
\end{figure*}
\para{Overview of Parameter Efficient Finetuning}
In contrast, we introduce a general framework of task-focused instruction tuning strategy, where the pretrained model AlpaCare~\cite{zhang2023alpacareinstructiontuned} is further finetuned for biomedical entity extraction task from clinical trial dataset.
In addition, the storage and deployment of finetuned models independently for each downstream task can incur significant expenses, given that fine-tuned models are of the same size as the original pretrained model. \hlt{Nevertheless, as the parameter sizes of large language models are huge, conducting comprehensive fine-tuning becomes infeasible due to the requirement of high resources. Also, it is infeasible to store them individually for different versions of frameworks produced after the task-specific finetuning~\cite{fu2022effectiveness} as the parameter sizes of the finetuned versions are the same as the non-finetuned version of frameworks.} To alleviate this problem, we employ the parameter efficient finetuning (PEFT) strategy by utilizing the low rank adaptation (LoRA)~\cite{lora} technique to finetune a small number of (extra) model parameters while freezing most parameters of the pretrained LLMs, thereby greatly decreasing the computational and storage costs. \hlt{LoRA applies a simple and efficient methodology to update the parameters of a weight matrix. This approach breaks down the higher dimensional matrix into a multiplication of two matrices with a low rank using the Kronecker product~\cite{krona}.} We consider LoRA as a reparametrization-based learning which can be formulated as follows:
\begin{align}
M_{o}=M_{i}W_{0}+M_{i}\Delta W=M_{i}W_{0}+ M_{i}BA,
\end{align}
where $W_0\in\mathrm{R}^{d\times d}$ is the pretrained weight matrix, including weights in the MLP or Attention layer.
$B\in\mathrm{R}^{r\times d}$ and $A\in\mathrm{R}^{r\times d}$ are lower-rank matrix intended for covering $\Delta W$ as depicted in Figure~\ref{fig:lora}. $r \ll d$ is an important hyper-parameter for LoRA. To conduct the instruction based finetuning for our downstream PICO frame extraction task, we prepare our own dataset in conversation style dataset.
\begin{figure*}[ht]%
    \centering
    \hspace*{-10.5mm}
    {\includegraphics[width=0.95\textwidth]{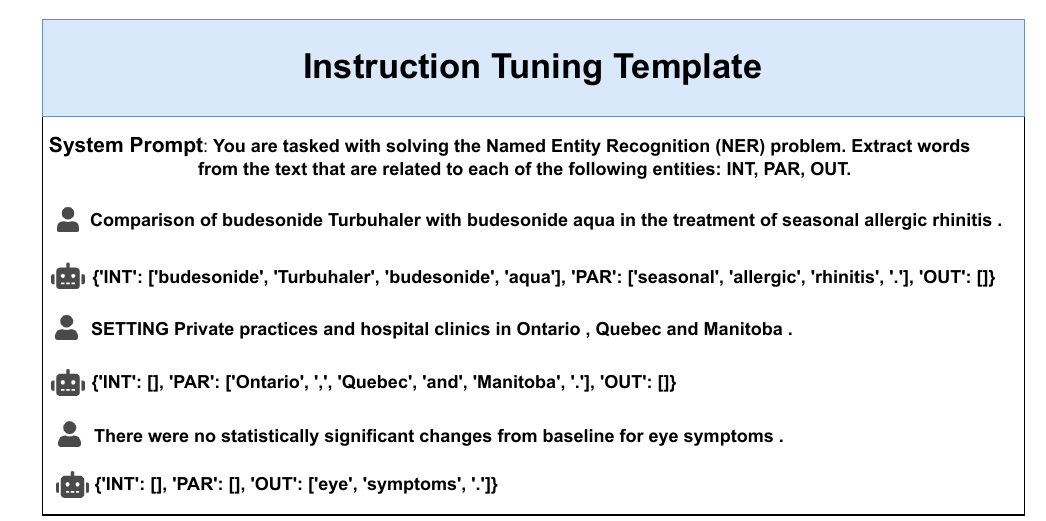}}
    % \captionsetup{width=0.7\textwidth}
     \vspace{-0.5em}
    \caption{\small The data construction prompt is utilized to produce entity mentions and their corresponding types for a specific passage.}
    \label{fig:inst_model}
\end{figure*}
As depicted in Figure~\ref{fig:inst_model}, we prepare our own conversation style dataset to conduct the instruction tuning approach for our downstream PICO frame extraction task. In general, most of the recent literature~\cite{zhou2023universalner} utilized state-of-the-art (SOTA) LLM such as ChatGPT~\cite{radford2019language} based framework to generate the task specific diverse samples to perform the instruction-tuning approach for our downstream task. In contrast to that, we utilize the existing annotated training dataset to prepare the conversation style dataset to conduct the downstream PICO frame extraction due to the limited resources. From Figure~\ref{fig:inst_model}, we can observe that our conversion style template which has three parts i.e., \texttt{instruction}, \texttt{input}, and \texttt{output}. In Figure~\ref{fig:inst_model} the `System Prompt' is used as the \texttt{instruction} whereas the \texttt{input} is used as sentences from the training data, and the \texttt{output} is the annotated entities present in that particular sentence. After obtaining conversation style annotated dataset, we propose a novel framework namely \textbf{AlpaPICO} by finetuning the \underline{Alpa}Care~\cite{zhang2023alpacare} in instruction-tuning~\cite{fang2023mol} based framework in an end-to-end fashion for our \underline{PICO} extraction task by employing LoRA technique.

\para{Implementation Details}
All the models were trained on Nvidia A100 80 GB GPU.
In terms of the common neural network settings, we used AdamW~\citep{loshchilov2018decoupled} as the optimizer with a learning rate of $0.0003$ and  a stopping criterion as mentioned in~\cite{conneau-etal-2020-unsupervised}; the training batch size used was 8.

\section{Experiment Setup}
In this section, we first describe the details of the datasets which have been used for our PICO frame extraction task from clinical trial literature and then we follow it up with the research questions to the task of PICO frame extraction, and the methods investigated towards addressing those questions.
\subsection{Dataset Description}
\label{ss:dataset}
\begin{table}[!t]
% \small
    % \centering
    \small
  \begin{center}
    \begin{adjustbox}{width=\columnwidth}
    \begin{tabular}{r|lll|ll}
    \hline
    &  & \multicolumn{2}{>{\columncolor{white}}l|}{Coarse-grained Dataset} & \multicolumn{2}{>{\columncolor{white}}l}{Fine-grained Dataset}\\
    \hline
      &\textbf{} & EBM-NLP & EBM-COMET & EBM-NLP$_{rev}$  & EBM-NLP$_{h}$\\
      \hline
      \multirow{2}{*}
      & \# total sentences & 53,397 & 5,193 & 40,092 & 53,404\\
      \hline
      \multirow{1}{*}{Training}
      & \# sentences & 40,935 & 3,895 & 30,069 & 40,942\\
      \hline
       \multirow{1}{*}{Validation}
       & \# sentences & 10,386 & 779 & 6,014 & 1,864\\
      \hline
      \multirow{1}{*}{Test}
      & \# sentences & 2,076 & 519  & 4,009 & 2,076\\
      \hline
    \end{tabular}
    \end{adjustbox}
  \end{center}
  % \vspace{-5mm}
    \caption{\small Dataset Statistics}
    \label{tab:datasets_stats}
 % \vspace{-3.0em}
\end{table}
To carry out our experimental investigations utilizing both our proposed ICL-based framework and the instruction-tuning-based method, we employ \hlt{on} four openly accessible datasets: EBMNLP, EBM-COMET, EBM-NLP\textsubscript{rev}, and EBM-NLP-hierarchical (EBM-NLP\textsubscript{h}). We have categorized all the datasets into two categories such as coarse-grained version and fine-grained version, according to its available annotations. More specifically speaking, entities belong to the given categories such as `Participation,' `Intervention,' or `Outcome' categories are considered as coarse-grained annotations. In contrast to that, the subdivision of a specific entity label into more detailed categories is considered as fine-grained annotation. It is important to note that, for fair comparison, during the evaluation phase, the fine-grained labels are mapped to coarse-grained labels. The fine-grained annotation is only utilized during the training phase. The Table~\ref{tab:datasets_stats} demonstrates the overall dataset statistics.
\begin{table}[!t]
\begin{adjustbox}{width=\columnwidth}
\small
\begin{tabular}{lll}
%  \hline
%        \textbf{EBM-NLP$_{h}$}\\
%             \hline
\hline
% \rowcolor{yellow}
\multicolumn{1}{>{\color{black}\columncolor{white}}l}{\textbf{Dataset Name}} & \multicolumn{1}{>{\color{black}\columncolor{white}}l}{\textbf{Coarse-grained Labels}} & \multicolumn{1}{>{\color{black}\columncolor{white}}l}{\textbf{Fine-grained Labels}}\\
\hline
\hline
\rowcolor{white}
& \mc{1}{\textbf{\textcolor{black}{Participants}}} & \mc{1}{\textcolor{black}{Age, Sex, Sample size, Condition}}\\
 % \rowcolor{green}
 \rowcolor{white}
\mc{1}{\textcolor{black}{\textbf{EBM-NLP$_{h}$}}} & \mc{1}{\textbf{\textcolor{black}{{Interventions}}}} & \mc{1}{\textcolor{black}{Surgical, Physical, Drug, Educational, Psychological, Control, Other}}  \\
\rowcolor{white}
& \mc{1}{\textbf{\textcolor{black}{Outcomes}}} & \mc{1}{\textcolor{black}{Physical, Pain, Mortality, Adverse effects, Mental, Other}} \\
 \hline
 % \rowcolor{aliceblue}
 % & \mc{1}{\textbf{Participants}} & \mc{1}{\textbf{\textemdash }}\\
 % \rowcolor{aliceblue}
 % \mc{1}{\textbf{ EBM-NLP$_{rev}$}} & \mc{1}{\textbf{Interventions}} &  \mc{1}{\textbf{\textemdash }}\\
 \rowcolor{white}
\mc{1}{\textcolor{black}{\textbf{\textcolor{black}{EBM-NLP$_{rev}$}}}} & \mc{1}{\textbf{\textcolor{black}{Outcomes}}} & \mc{1}{\textcolor{black}{Physical, Pain, Mortality, Adverse effects, Mental, Other}} \\
 \hline
  \hline
\end{tabular}
\end{adjustbox}
\vspace{0.8em}
\caption{\small Description of fine-grained entity types present in EBM-NLP$_{rev}$ and EBM-NLP-hierarchical (EBM-NLP$_{h}$).}
\label{tab:fine_grained_stats}
% \vspace{-3.2em}
\end{table}
The Table~\ref{tab:fine_grained_stats} demonstrates that in the EBM-NLP$_{h}$ dataset, each coarse-grained category is further divided into fine-grained labels, while the EBM-NLP$_{rev}$ contains finer labels only for one broader label.
\subsection{Research Questions}
In this section we pose a few important research questions, which are central to our research work.
\para{RQ-1}
What is the feasibility of applying in-context learning based framework for our downstream PICO frame extraction task from clinical trial text using AlpaCare language model?
\para{RQ-2}
How effectively can we apply instruction tuning-based techniques using an LLM in extracting the PICO related terminologies for both coarse-grained and fine-grained variants of the dataset?

\subsection{Methods Investigated}

\subsubsection{Baselines for Instruction Tuning Setup}
\label{ss:baseline_instruction}
The PICO frame extraction task in clinical trial literature is commonly viewed as a sequence labeling task. Unlike traditional supervised frameworks, our proposed novel AlpaPICO considers the PICO frame extraction as a sequence generation task. By evaluating its performance against state-of-the-art supervised frameworks, we demonstrate the efficacy of AlpaPICO, emphasizing its effectiveness in addressing the challenges of natural language generation tasks compared to sequence classification tasks~\cite{yang-etal-2018-sgm}.
\begin{itemize}
    \item BioBERT~\cite{lee2020biobert}:
    In particular, Lee et.al.~\cite{lee2020biobert} employed Bio-BERT to fine-tune it on the downstream PICO frame extraction to leverage the pretrained biomedical knowledge of the Bio-BERT, acquired during the pretraining phase by the BioBERT language model.
    \item SciBERT~\cite{beltagy-etal-2019-scibert}: Similar to BioBERT work, we finetune the SciBERT language model for the downstream the PICO frame extraction task by considering it as sequence labeling task. 
    \item BioLinkBERT-Large~\cite{yasunaga-etal-2022-linkbert}: Additionally, in order to show the efficacy of our proposed novel AlpaPICO framework, we also apply finetuning strategy on BioLinkBERT language for our downstream PICO frame extraction task. 
    \item Llama-2-Inst~\cite{touvron2023llama}: We finetune the Llama-2 large language model by converting our PICO frame extraction sequence classification dataset to conversation style format as described in Section~\ref{ss:inst_tuning}.
\end{itemize}

\subsubsection{Baselines for In-context Learning Setup}
To show the effectiveness of our proposed in-context learning based framework we compare the performance with the following baselines. Although ICL-based framework eliminates the conventional training process, we compare its performance with the purely supervised frameworks to leverage the effectiveness of pretrained knowledge of LLM, gathered during the pretraining phase. 
\begin{itemize}
    \item Zero-Shot: In this configuration, we instruct our language model AlpaCare without supplying any additional annotated context i.e., annotated demonstrations as described in Section~\ref{ss:icl_framework} to perform the PICO frame extraction task by considering the sequence generation task.
    \item BioLinkBERT-Large~\cite{yasunaga-etal-2022-linkbert}: In order to achieve optimal performance, particularly when an annotated training dataset is accessible for training purposes, we undertake fine-tuning of the BioLinkBERT language model specifically for the downstream PICO frame extraction task.
\end{itemize}

\subsubsection{Result \& Analysis}
\label{ss:results}
\begin{table}[!t]
\centering
\small
\begin{tabular}{@{}l@{~~~}l c|c|c|c@{}}
\toprule
\multirow{2}{*}{Model} & \multirow{2}{*}{} &
\multicolumn{1}{c|}{EBM-NLP} & \multicolumn{1}{c|}{EBM-NLP$_{h}$} & \multicolumn{1}{c|}{EBM-NLP$_{rev}$} & \multicolumn{1}{c}{EBM-COMET} \\
\cmidrule(r){3-6}

& & F-score & F-score & F-score & F-score \\
\midrule
\multirow{2}{*}{\rotatebox{90}{Baselines}}
& Zero-shot (ICL$_0$) & 0.020 & 0.048 & 0.00 & 0.00\\
& BioLinkBERT-Large~\cite{yasunaga-etal-2022-linkbert} & \textbf{ 74.19} & \underline{43.95} & \textbf{80.41} & \textbf{62.52}\\
% & AlpaPICO & 64.81 & 62.33 & 70.90 & 70.12\\
\midrule
\multirow{1}{*}{\rotatebox{90}{ICL}}
& $k$-shot ICL & \underline{47.21} & \textbf{ 63.10} & \underline{40.59} & \underline{43.74}\\
\bottomrule
\end{tabular}
\caption{\small A comparison between \textbf{F-scores} obtained from supervised PLMs, their respective in-domain (biomedical) versions, and our instruction tuning based approach \textbf{AlpaPICO}. The best results have been shown in \textbf{bold} and the second best results have been underlined. Values of $k$ are 3, 4, 9, 9 for EBM-NLP, EBM-NLP\textsubscript{h}, EBM-NLP$_{rev}$, EBM-COMET   respectively.
}
\label{tab:result_icl}
\end{table}

\begin{table}[h!]
\centering
\small
\begin{tabular}{l| c| c| c |c}
\toprule
% \rowcolor{blue}
Dataset & Precision &  Recall & F-Score & Accuracy\\
\midrule
\midrule
EBM-NLP$_{rev}$ & 48.08 & 48.35 & 47.21 & 31.15\\ 
EBM-COMET & 45.06 & 36.93 & 40.59 & 25.47\\
\bottomrule
\bottomrule
\end{tabular}
\caption{\hlt{The overall performance of our $k$-shot ICL based technique using a pre-trained LLM \textbf{AlpaPICO}.} \hlt{Values of $k$ are 3, 4, 9, 9 for EBM-NLP$_{rev}$, and EBM-COMET respectively.}} 
\label{tab:Alpapico_ICL}
\end{table}

\begin{table}[ht]
\centering
\small
\begin{tabular}{l| c c c  c || c c c c }
\toprule
\toprule
\textbf{} & \multicolumn{4}{c||}{EBM-NLP} & \multicolumn{4}{c}{EBM-NLP$_{h}$} \\ \hline
 & Precision & Recall & F-score & Accuracy & Precision & Recall & F-score & Accuracy \\ \hline
OUT & 49.46 & 35.59 & 41.40 & 26.10 & 62.02 & 49.26 & 54.91 & 37.84 \\ 
% \hline
INT & 36.47 & 54.71 & 43.76 & 28.01 & 67.76 & 52.56 & 59.20 & 42.05\\ 
% \hline
PAR & 58.32 & 54.74 & 56.47 & 39.35 & 79.39 & 71.42 & 75.19 & 60.25 \\ 
% \hline
\bottomrule
\bottomrule
\end{tabular}
\caption{\hlt{Class-wise performace of our $k$-shot ICL based technique using a pre-trained LLM \textbf{AlpaPICO}.} \hlt{Values of $k$ are 3,4,9,9 for EBM-NLP and EBM-NLP$_{h}$ respectively.}}
\label{tab:cg_fg_icl}
\end{table}
\begin{figure*}[!t]%
    \centering
    % \hspace*{-10.5mm}
    {\includegraphics[width=0.45\textwidth]{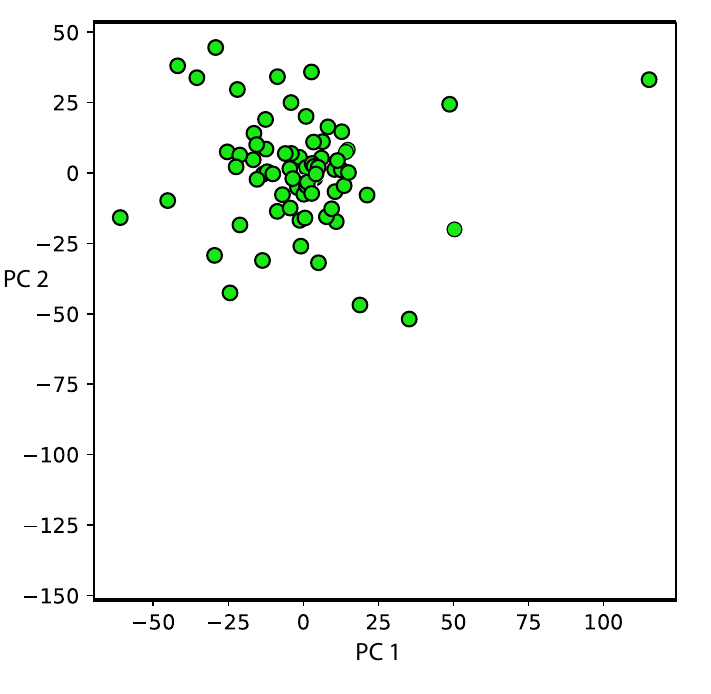}}
    % \captionsetup{width=0.7\textwidth}
     \vspace{-0.5em}
    \caption{\small Dimensionally-reduced feature vector representation of the randomly sampled 1000 training instances from EBM-NLP. Later on, these vectors underwent additional reduction to only two dimensions using PCA for the purpose of visualization. Note that the diversity in the training instances of EBM-NLP dataset is less.}
    \label{fig:diverse}
     % \vspace{-1.5em}
\end{figure*}
To investigate RQ-1, Table~\ref{tab:result_icl} shows a comparison between the methods investigated for PICO frame extraction task. \hlt{
Table~\ref{tab:Alpapico_ICL} shows performance of $k$-shot ICL based technique for EBM-NLP$_{rev}$, and EBM-COMET. Also, Table~\ref{tab:cg_fg_icl} shows the overall results for $k$-shot ICL for EBM-NLP and EBM-NLP$_{h}$ respectively.} It can be seen that task-specific finetuning of any language model produces the best results (as seen for BioLinkBERT-Large). However, a supervised framework requires the existence of a training set of labeled examples of sentences and the annotated entities present in the sentence. In contrast to that, even a completely unsupervised zero-shot scenario ICL$_0$ and without any training set yields results that are substantially worse in terms of F-score than the supervised frameworks. This observation leads to further investigation of whether the addition of annotated context with the system prompt to instruct the AlpaCare framework for the downstream PICO frame extraction task. We can observe from Table~\ref{tab:result_icl} that the addition of annotated context along with the task description as described in Figure~\ref{fig:io} helps significantly to improve the performance of AlpaCare~\footnote{\url{xz97/AlpaCare-llama2-7b}} LLM for our downstream PICO frame extraction task in an unsupervised setup. More specifically speaking, our proposed $k$-shot ICL-based framework, where $k$ represents the number of annotated instances formally referred to as demonstrations (cf. Section~\ref{ss:icl_framework}), performs well across various datasets. The probable explanation for this phenomenon is that when AlpaCare, finetuned on biomedical data instances, encounters the $k$ number of semantically similar instances from the training dataset, it effectively leverages its pretrained knowledge, acquired during the pretraining phase. This enables AlpaCare to generate relevant entities more adeptly in comparison to the zero-shot scenario, where no additional demonstrations are provided alongside the task description. Additionally, we can also observe from Table~\ref{tab:result_icl} that the performance of our proposed $k$-shot ICL framework is relatively comparable to the supervised framework in most of the datasets such as EBM-NLP, EBM-NLP$_{rev}$ and EBM-COMET whereas our $k$-shot ICL based framework outperforms all the supervised frameworks on EBM-NLP$_{h}$ dataset. The rationale underlying this observed performance is rooted in the fact that the ICL-based framework treats our downstream PICO frame extraction task as a natural language generation task, diverging from the conventional sequence classification task employed, for instance, finetuning BioLinkBERT-Large on the PICO frame extraction from the clinical trial dataset. Natural language generation inherently poses greater challenges, as highlighted in the work by Yang et al.~\cite{yang-etal-2018-sgm}, compared to classification problems. Additionally, another key insight is that the ICL framework tends to yield optimal performance when presented with a diverse set of annotated instances. In our specific case, analysis of the embedding space, as depicted in
Figure~\ref{fig:diverse}, reveals that a substantial portion of training instances from the EBM-NLP dataset forms a cohesive cluster, suggesting the importance of diversity in annotated instances for achieving satisfactory performance. An interesting observation concerning the fine-grained EBM-NLP$_{h}$ dataset, as mentioned in Table~\ref{tab:result_icl}, is that our $k$-shot ICL-based framework outperforms both the zero-shot and supervised BioLink-BERT-Large framework in the PICO frame extraction task. The likely reason is that the clean and fine-grained annotations present in the EBM-NLP$_{h}$ dataset, as compared to the EBM-NLP, EBM-COMET, and EBM-NLP$_{rev}$ datasets where clean annotated demonstrations help our ICL framework to outperform the baseline models in terms of F-score.
\begin{table}[!t]
%\vspace{-.5cm}
\centering
%\small
\begin{adjustbox}{width=.75\textwidth}

%\begin{tabular}{c|ccc|ccc}
\small
\begin{tabular}{l| c| c| c |c}
\toprule
Model &EBM-NLP & EBM-NLP$_{rev}$ & EBM-COMET & EBM-NLP$_{h}$\\
\midrule
\midrule
BioBERT~\cite{lee2020biobert} & 73.18 & 53.10 & \textbf{81.50} & --\\
SciBERT~\cite{beltagy-etal-2019-scibert} & 73.06 & 52.80 & 77.60 & --\\
BioLinkBERT-Large~\cite{yasunaga-etal-2022-linkbert} & \textbf{74.19} & 43.95 & 80.41 & 62.52\\

Llama-2-Inst. & 60.51 & 59.36 & 64.48 & 69.86 \\

AlpaPICO & \underline{64.81} & \textbf{62.33} & \underline{70.90} & \textbf{70.12}\\
\bottomrule
\bottomrule
\end{tabular}
\end{adjustbox}
\vspace{0.7em}
\caption{\small A comparison between \textbf{F-scores} obtained from supervised PLMs, their respective in-domain (biomedical) versions, and our instruction tuning based approach \textbf{AlpaPICO}. The best results have been shown in \textbf{bold} and the second best results have been underlined.
}
\label{tab:result_instruct}
\end{table}
\begin{figure*}[!t]%
    \centering
    % \hspace*{-10.5mm}
    {\includegraphics[width=0.65\textwidth]{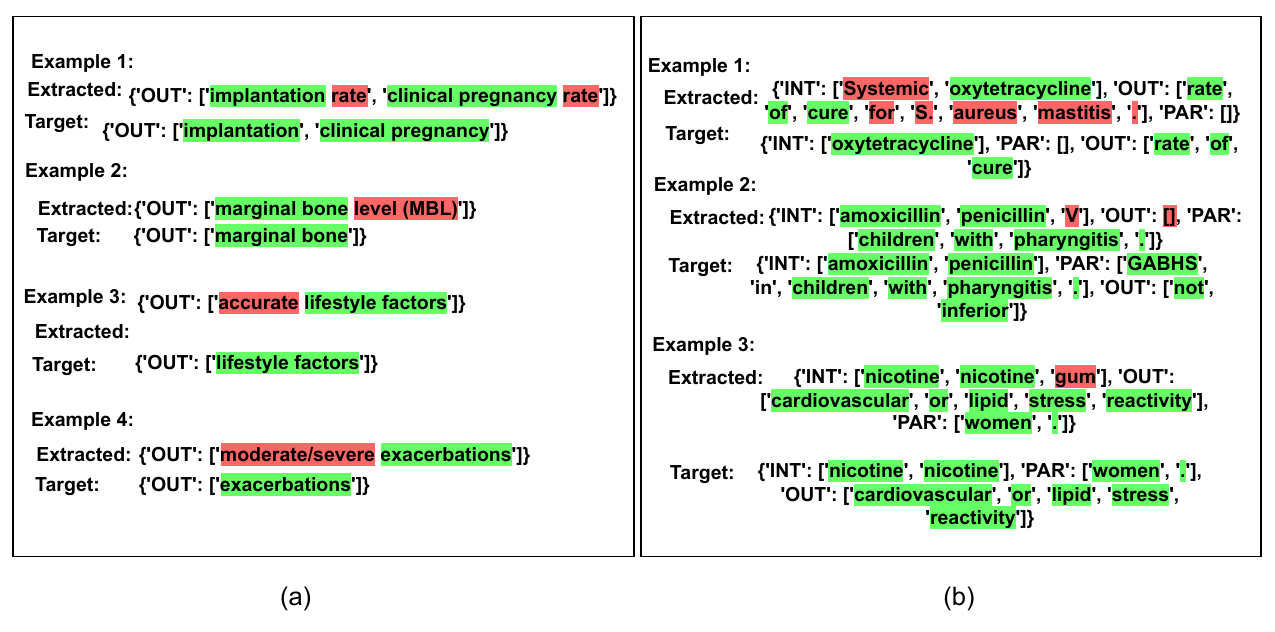}}
    % \captionsetup{width=0.7\textwidth}
     \vspace{-0.5em}
    \caption{\small Comparison of PICO frame entities generated by our proposed instruction tuned AlpaPICO framework and the actual annoated entities on coarse-grained (a) EBM-COMET and (b) EBM-NLP datasets.}
    \label{fig:comet_ebm}
     % \vspace{-1.5em}
\end{figure*}
\begin{table}[h!]
\centering
\small
\begin{tabular}{l| c| c| c |c}
\toprule
% \rowcolor{blue}
Dataset & Precision &  Recall & F-Score & Accuracy\\
\midrule
\midrule
EBM-NLP$_{rev}$ & 85.15 & 49.16 & 62.33 & 45.27\\ 
% \hline
EBM-COMET & 81.40 & 62.80 & 70.90 & 54.90\\
% EBM-NLP$_{h}$ & 75.10 & 66.00 & 70.12          & 54.30 \\
% \hline
\bottomrule
\bottomrule
\end{tabular}
\caption{\hlt{Performance of our instruction based approach \textbf{AlpaPICO}}}
\label{tab:Alpapico_new_stats}
\end{table}
\begin{table}[ht]
\centering
\small
\begin{tabular}{l| c c c  c || c c c c }
\toprule
\toprule
\textbf{} & \multicolumn{4}{c||}{EBM-NLP} & \multicolumn{4}{c}{EBM-NLP$_{h}$} \\ \hline
 & Precision & Recall & F-score & Accuracy & Precision & Recall & F-score & Accuracy \\ \hline
OUT & 85.88 & 49.03 & 62.42 & 45.37 & 65.87 & 63.66 & 64.75 & 47.87 \\ 
% \hline
INT & 64.21 & 49.91 & 56.17 & 39.05 & 78.16 & 59.54 & 67.59 & 51.05 \\ 
% \hline
PAR & 82.38 & 70.28 & 75.85 & 61.09 & 81.4 & 74.91 & 78.02 & 63.97 \\ 
% \hline
\bottomrule
\bottomrule
\end{tabular}
\caption{\hlt{Class-wise performance of our instruction tuning based approach \textbf{AlpaPICO} on EBM-NLP and EBM-NLP$_{h}$ datasets.}}
\label{tab:cg_fg_instruct}
\end{table}

To explore RQ-2, we perform an instruction tuning based technique using AlpaCare LLM on both coarse-grained and fine-grained datasets. We evaluate the performance of our proposed AlpaPICO framework in terms of F-score. Table~\ref{tab:result_instruct} shows an interesting sets of observations, which presents that our proposed instruction tuning chat model AlpaPICO produces comparable results on coarse-grained datasets such as EBM-NLP and EBM-COMET datasets. \hlt{Table~\ref{tab:Alpapico_new_stats} and Table~\ref{tab:cg_fg_instruct} show detailed observations of \textbf{AlpaPICO} using our instruction based tuning based approach EBM-NLP, EBM-NLP$_{h}$, EBM-NLP$_{rev}$ and EBM-COMET datasets respectively.} The rationale behind this performance is that the annotations of both the coarse-grained EBM-NLP and EBM-COMET datasets are noisy in nature. Additionally, from Figure~\ref{fig:comet_ebm}, we can observe that the generated entities for the first test instance of EBM-COMET dataset is `implantation rate' and `clinical pregnancy rate' whereas the actual annotation contains the `implantation' and `clinical pregnancy' respectively. Another example of the generated entities of the test instance of the EBM-NLP dataset is `systematic oxytretracycline' whereas the actual annotation is `oxytretracycline'. Hence, from both example snippets, it is evident that our proposed AlpaPICO framework can successfully generate central entity tokens, while also generating additional tokens due to the presence of noisy annotations in the training instances. However, since we employ a token-level strict matching technique to calculate the F-score, our AlpaPICO framework does not outperform the performance of existing supervised baselines on these two datasets. In contrast to the aforementioned observation, another noteworthy finding is that our instruction-tuned AlpaPICO framework exhibits a significant performance advantage over existing supervised baselines in terms of F-score on both the fine-grained EBM-NLP$_{rev}$ and EBM-NLP$_h$ datasets. The likely reason behind this observation is that the combination of fine-grained and clear annotations, as detailed in Section~\ref{ss:dataset}, contributes substantially to enhancing the performance of our instruction-tuned AlpaPICO framework. 

\section{Ablation Study}
\begin{figure*}[!t]%
    \centering
    \hspace*{-10.5mm}
    {\includegraphics[width=0.35\textwidth]{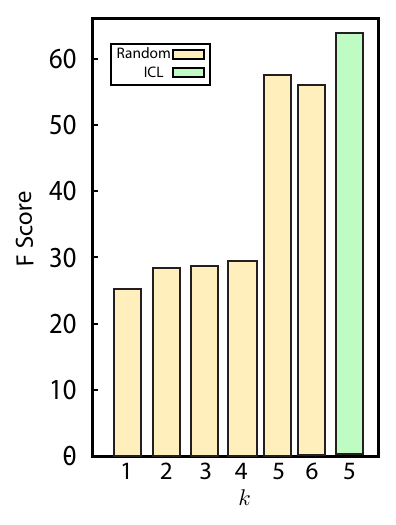}}
    % \captionsetup{width=0.7\textwidth}
     \vspace{-0.5em}
    \caption{\small Performance (F-score on Y-axis) comparison based on random context selection, by changing the number of samples ($k$ on X-axis), in the ICL-based framework.}
    \label{fig:icl_rand}
     % \vspace{-1.5em}
\end{figure*}

Throughout the experiments with our $k$-shot ICL framework, we employ two strategies to select the relevant annotated contexts, which are considered as demonstrations, aiming to enhance the performance of our proposed ICL framework. In selecting the relevant annotated context, we employ a strategy based on dense index and cosine similarity. Specifically, we acquire the BioLinkBERT embedding of a particular test instance from the EBM-NLP dataset, where the dense embedding of the training instances is pre-indexed using the FAISS framework\footnote{\url{https://github.com/facebookresearch/faiss}}. The test instance embedding serves as the necessary query embedding, which is passed to the FAISS dense indexer. This indexer utilizes the hierarchical navigable small worlds (HNSW)~\cite{malkov2018efficient} based nearest neighbor searching strategy to retrieve semantically relevant instances. In contrast, we also implement a random selection strategy to retrieve the annotated context from the training dataset corresponding to a specific test instance. It is noteworthy as depicted in Figure~\ref{fig:icl_rand}, that our randomly selected $5$ examples from EBM-NLP training dataset achieves highest F-score but it cannot outperform the best-performing $5$-shot ICL, where the required context is selected using the HNSW algorithm. So, we perform additional experiments by employing ICL framework, selecting semantically similar contexts corresponding to specific test instances across the remaining datasets. 

\begin{figure*}[!t]%
    \centering
    \hspace*{-10.5mm}
    {\includegraphics[width=0.70\textwidth]{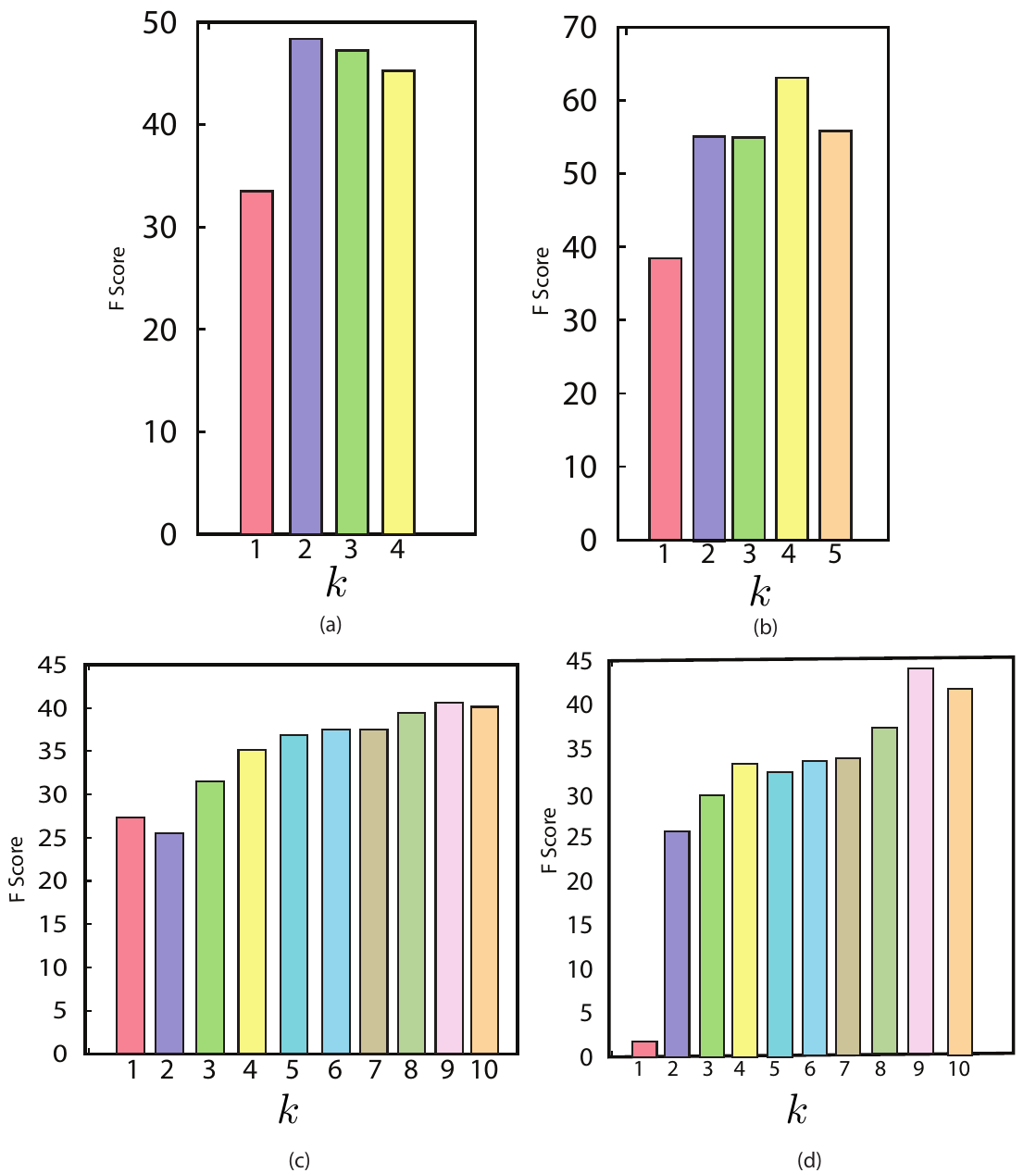}}
    % \captionsetup{width=0.7\textwidth}
     \vspace{-0.5em}
    \caption{\small Performance (F-score on Y-axis) comparison of our $k$-shot ICL framework on different datasets (a) EBM-NLP, (b) EBM-NLP\textsubscript{h}, (c) EBM-NLP\textsubscript{rev}, (d) EBM-COMET by changing the number of context ($k$).}
    \label{fig:ablation_fig}
     % \vspace{-1.5em}
\end{figure*}
However, we carry out more ablation experiments for our $k$-shot ICL framework to select the optimized $k$ value for our downstream PICO frame extraction task from the clinical trial literature. From Figure~\ref{fig:ablation_fig}, we observe that the optimized $k$-values are not the same for all the datasets. 
\hlt{For instance, we attain the highest F-score on the EBM-NLP dataset within the ICL framework for $k$ = $3$. Conversely, for the EBM-NLP$_{h}$, EBM-NLP$_{rev}$, and EBM-COMET datasets, the $k$ values have been set to $4$, $9$, and $9$, respectively.} The probable reason for this phenomenon is that the ICL-based framework is inherently responsive and context-sensitive~\cite{wuself}.
\section{Concluding Remarks}
In this work, we investigate the feasibility of applying In-context learning framework as well as instruction tuning approach for PICO frame extraction task from clinical trial documents. Our novel $k$-shot ICL based framework for PICO frame extraction task in the evidence-based medicine literature perform significantly well, without applying any training operation. \hlt{Additionally, we propose a supervised instruction tuning based framework in low resource environment, namely AlpaPICO, which produces the state-of-the-art (SOTA) performance on both EBM-NLP$_{rev}$ and EBM-NLP$_{h}$ datasets. It also produces comparable results on the two remaining datasets. As both of our approaches consider the PICO frame extraction as a natural language generation task instead of a sequence classification task, our smaller case-study, as depicted in Figure~\ref{fig:comet_ebm} shows the effectiveness of LLM towards generating the PICO frame extraction task. A limitation of our work is that the computation is memory-intensive due to the use of LLMs, thus making it difficult to execute our approach on a terminal with limited memory capacity.} 
Additionally, due to the limitations of our own resources, we could not use a larger or a commercially accessible LLM. In future, we plan to improve the performance of ICL framework by selecting the relevant context from external corpus, such as~\textbf{Cochrane} database. We can also leverage commercial Large Language Models (LLMs) to generate state-of-the-art data instances for evidence-based medicine literature. These instances can be used to apply knowledge distillation processes on smaller versions of LLMs for the PICO frame extraction task from clinical trial documents.

\section*{Author Contribution:}
MG, SM, PBC, SKN and DG conceptualized the study. MG, SM and AG performed experimental work. MG and SM contributed to analysis the results and prepare the figures. MG, SM and AG wrote the initial draft of the manuscript. MG, SM, PBC, SKN, DG edited the manuscript.  
\section*{Conflicts of interest:}
The authors declare that there are no conflicts of interest.
\section*{Funding information:}
This study was funded by Indian Association for the Cultivation of Science (IACS), Kolkata, India.

% \label{}







\bibliographystyle{elsarticle-num}
\bibliography{ref}

\begin{thebibliography}{100}
\expandafter\ifx\csname url\endcsname\relax
  \def\url#1{\texttt{#1}}\fi
\expandafter\ifx\csname urlprefix\endcsname\relax\def\urlprefix{URL }\fi
\expandafter\ifx\csname href\endcsname\relax
  \def\href#1#2{#2} \def\path#1{#1}\fi

\bibitem{sackett1997evidence}
D.~L. Sackett, Evidence-based medicine, in: Seminars in perinatology, Vol.~21, Elsevier, 1997, pp. 3--5.

\bibitem{cook1997systematic}
D.~J. Cook, C.~D. Mulrow, R.~B. Haynes, Systematic reviews: synthesis of best evidence for clinical decisions, Annals of internal medicine 126~(5) (1997) 376--380.

\bibitem{jonnalagadda2015automating}
S.~R. Jonnalagadda, P.~Goyal, M.~D. Huffman, Automating data extraction in systematic reviews: a systematic review, Systematic reviews 4~(1) (2015) 1--16.

\bibitem{boudin2010positional}
F.~Boudin, J.-Y. Nie, M.~Dawes, Positional language models for clinical information retrieval, in: Proceedings of the 2010 Conference on Empirical Methods in Natural Language Processing, 2010, pp. 108--115.

\bibitem{marshall-etal-2017-automating}
I.~Marshall, J.~Kuiper, E.~Banner, B.~C. Wallace, \href{https://aclanthology.org/P17-4002}{Automating biomedical evidence synthesis: {R}obot{R}eviewer}, in: M.~Bansal, H.~Ji (Eds.), Proceedings of {ACL} 2017, System Demonstrations, Association for Computational Linguistics, Vancouver, Canada, 2017, pp. 7--12.
\newline\urlprefix\url{https://aclanthology.org/P17-4002}

\bibitem{nye-etal-2018-corpus}
B.~Nye, J.~J. Li, R.~Patel, Y.~Yang, I.~Marshall, A.~Nenkova, B.~Wallace, A corpus with multi-level annotations of patients, interventions and outcomes to support language processing for medical literature, ACL, Melbourne, Australia, 2018.

\bibitem{jin2018pico}
D.~Jin, P.~Szolovits, Pico element detection in medical text via long short-term memory neural networks, in: Proceedings of the BioNLP 2018 workshop, 2018, pp. 67--75.

\bibitem{kim2011automatic}
S.~N. Kim, D.~Martinez, L.~Cavedon, L.~Yencken, Automatic classification of sentences to support evidence based medicine, in: BMC bioinformatics, Vol.~12, BioMed Central, 2011, pp. 1--10.

\bibitem{huang2006evaluation}
X.~Huang, J.~Lin, D.~Demner-Fushman, Evaluation of pico as a knowledge representation for clinical questions, in: AMIA, Vol. 2006, American Medical Informatics Association, 2006, p. 359.

\bibitem{abaho2019correcting}
M.~Abaho, D.~Bollegala, P.~Williamson, S.~Dodd, Correcting crowdsourced annotations to improve detection of outcome types in evidence based medicine, in: CEUR Workshop Proceedings, Vol. 2429, 2019, pp. 1--5.

\bibitem{abaho2022assessment}
M.~Abaho, D.~Bollegala, P.~R. Williamson, S.~Dodd, Assessment of contextualised representations in detecting outcome phrases in clinical trials, arXiv preprint arXiv:2203.03547 (2022).

\bibitem{stubbs2015automated}
A.~Stubbs, C.~Kotfila, {\"O}.~Uzuner, Automated systems for the de-identification of longitudinal clinical narratives: Overview of 2014 i2b2/uthealth shared task track 1, Journal of biomedical informatics 58 (2015) S11--S19.

\bibitem{uzuner2007evaluating}
{\"O}.~Uzuner, Y.~Luo, P.~Szolovits, Evaluating the state-of-the-art in automatic de-identification, Journal of the American Medical Informatics Association 14~(5) (2007) 550--563.

\bibitem{wu2023pmcllama}
C.~Wu, W.~Lin, X.~Zhang, Y.~Zhang, Y.~Wang, W.~Xie, Pmc-llama: Towards building open-source language models for medicine (2023).
\newblock \href {http://arxiv.org/abs/2304.14454} {\path{arXiv:2304.14454}}.

\bibitem{luo2023biomedgpt}
Y.~Luo, J.~Zhang, S.~Fan, K.~Yang, Y.~Wu, M.~Qiao, Z.~Nie, Biomedgpt: Open multimodal generative pre-trained transformer for biomedicine (2023).
\newblock \href {http://arxiv.org/abs/2308.09442} {\path{arXiv:2308.09442}}.

\bibitem{zhang2023alpacare}
X.~Zhang, C.~Tian, X.~Yang, L.~Chen, Z.~Li, L.~R. Petzold, Alpacare: Instruction-tuned large language models for medical application, arXiv preprint arXiv:2310.14558 (2023).

\bibitem{beltagy-etal-2019-scibert}
I.~Beltagy, K.~Lo, A.~Cohan, {S}ci{BERT}: A pretrained language model for scientific text, in: Proceedings of the 2019 EMNLP, ACL, Hong Kong, China, 2019, pp. 3615--3620.

\bibitem{yasunaga-etal-2022-linkbert}
M.~Yasunaga, J.~Leskovec, P.~Liang, {L}ink{BERT}: Pretraining language models with document links, in: Proceedings of the 60th Annual Meeting of the ACL (Volume 1: Long Papers), ACL, Dublin, Ireland, 2022, pp. 8003--8016.

\bibitem{hoory2021learning}
S.~Hoory, A.~Feder, A.~Tendler, A.~Cohen, S.~Erell, I.~Laish, H.~Nakhost, U.~Stemmer, A.~Benjamini, A.~Hassidim, Y.~Matias, Learning and evaluating a differentially private pre-trained language model, in: Proceedings of the Third Workshop on Privacy in Natural Language Processing, ACL, Online, 2021, pp. 21--29.

\bibitem{mayhew2020robust}
S.~Mayhew, G.~Nitish, D.~Roth, Robust named entity recognition with truecasing pretraining, in: Proceedings of the AAAI Conference on Artificial Intelligence, Vol.~34, 2020, pp. 8480--8487.

\bibitem{sung2021cnnbif}
C.~Sung, V.~Goel, E.~Marcheret, S.~Rennie, D.~Nahamoo, {CNNB}i{F}: {CNN}-based bigram features for named entity recognition, ACL, 2021.

\bibitem{li2020attention}
P.-H. Li, T.-J. Fu, W.-Y. Ma, Why attention? analyze bilstm deficiency and its remedies in the case of ner, in: Proceedings of the AAAI Conference on Artificial Intelligence, Vol.~34, 2020, pp. 8236--8244.

\bibitem{chen2019grn}
H.~Chen, Z.~Lin, G.~Ding, J.~Lou, Y.~Zhang, B.~Karlsson, Grn: Gated relation network to enhance convolutional neural network for named entity recognition, in: Proceedings of the AAAI Conference on Artificial Intelligence, Vol.~33, 2019, pp. 6236--6243.

\bibitem{xu2021supervised}
Y.~Xu, H.~Huang, C.~Feng, Y.~Hu, A supervised multi-head self-attention network for nested named entity recognition, in: Proceedings of the AAAI Conference on Artificial Intelligence, Vol.~35, 2021, pp. 14185--14193.

\bibitem{li2020recursively}
B.~Li, S.~Liu, Y.~Sun, W.~Wang, X.~Zhao, Recursively binary modification model for nested named entity recognition, in: Proceedings of the AAAI Conference on Artificial Intelligence, Vol.~34, 2020, pp. 8164--8171.

\bibitem{dai2019joint}
D.~Dai, X.~Xiao, Y.~Lyu, S.~Dou, Q.~She, H.~Wang, Joint extraction of entities and overlapping relations using position-attentive sequence labeling, in: Proceedings of the AAAI conference on artificial intelligence, Vol.~33, 2019, pp. 6300--6308.

\bibitem{zeng2020copymtl}
D.~Zeng, H.~Zhang, Q.~Liu, Copymtl: Copy mechanism for joint extraction of entities and relations with multi-task learning, in: Proceedings of the AAAI conference on artificial intelligence, Vol.~34, 2020, pp. 9507--9514.

\bibitem{nayak2020effective}
T.~Nayak, H.~T. Ng, Effective modeling of encoder-decoder architecture for joint entity and relation extraction, in: Proceedings of the AAAI conference on artificial intelligence, Vol.~34, 2020, pp. 8528--8535.

\bibitem{xiao2020joint}
Y.~Xiao, C.~Tan, Z.~Fan, Q.~Xu, W.~Zhu, Joint entity and relation extraction with a hybrid transformer and reinforcement learning based model, in: Proceedings of the AAAI Conference on Artificial Intelligence, Vol.~34, 2020, pp. 9314--9321.

\bibitem{sun2021progressive}
K.~Sun, R.~Zhang, S.~Mensah, Y.~Mao, X.~Liu, Progressive multitask learning with controlled information flow for joint entity and relation extraction, Association for the Advancement of Artificial Intelligence (AAAI) (2021).

\bibitem{li2021joint}
R.~Li, D.~Li, J.~Yang, F.~Xiang, H.~Ren, S.~Jiang, L.~Zhang, Joint extraction of entities and relations via an entity correlated attention neural model, Information Sciences 581 (2021) 179--193.

\bibitem{ji-etal-2021-neural}
Z.~Ji, T.~Xia, M.~Han, J.~Xiao, \href{https://aclanthology.org/2021.acl-long.219}{A neural transition-based joint model for disease named entity recognition and normalization}, in: Proceedings of the 59th Annual Meeting of the Association for Computational Linguistics and the 11th International Joint Conference on Natural Language Processing (Volume 1: Long Papers), Association for Computational Linguistics, Online, 2021, pp. 2819--2827.
\newblock \href {https://doi.org/10.18653/v1/2021.acl-long.219} {\path{doi:10.18653/v1/2021.acl-long.219}}.
\newline\urlprefix\url{https://aclanthology.org/2021.acl-long.219}

\bibitem{DBLP:journals/talip/DasGG17}
A.~Das, D.~Ganguly, U.~Garain, \href{https://doi.org/10.1145/3015467}{Named entity recognition with word embeddings and wikipedia categories for a low-resource language}, {ACM} Trans. Asian Low Resour. Lang. Inf. Process. 16~(3) (2017) 18:1--18:19.
\newblock \href {https://doi.org/10.1145/3015467} {\path{doi:10.1145/3015467}}.
\newline\urlprefix\url{https://doi.org/10.1145/3015467}

\bibitem{mukherjee-etal-2023-mllab4cs}
S.~Mukherjee, M.~Ghosh, {Girish}, P.~Basuchowdhuri, \href{https://aclanthology.org/2023.semeval-1.192}{{ML}lab4{CS} at {S}em{E}val-2023 task 2: Named entity recognition in low-resource language {B}angla using multilingual language models}, in: A.~K. Ojha, A.~S. Do{\u{g}}ru{\"o}z, G.~Da~San~Martino, H.~Tayyar~Madabushi, R.~Kumar, E.~Sartori (Eds.), Proceedings of the 17th International Workshop on Semantic Evaluation (SemEval-2023), Association for Computational Linguistics, Toronto, Canada, 2023, pp. 1388--1394.
\newblock \href {https://doi.org/10.18653/v1/2023.semeval-1.192} {\path{doi:10.18653/v1/2023.semeval-1.192}}.
\newline\urlprefix\url{https://aclanthology.org/2023.semeval-1.192}

\bibitem{wang2021learning}
D.~Wang, H.~Fan, J.~Liu, Learning with joint cross-document information via multi-task learning for named entity recognition, Information Sciences 579 (2021) 454--467.

\bibitem{ma2023sequence}
Q.~Ma, L.~Yu, H.~Chen, J.~Yan, Z.~Lin, Sequence labeling with mlta: Multi-level topic-aware mechanism, Information Sciences 637 (2023) 118934.

\bibitem{toledo2019information}
J.~I. Toledo, M.~Carbonell, A.~Forn{\'e}s, J.~Llad{\'o}s, Information extraction from historical handwritten document images with a context-aware neural model, Pattern Recognition 86 (2019) 27--36.

\bibitem{ghosh-etal-2022-astro}
M.~Ghosh, P.~Santra, S.~A. Iqbal, P.~Basuchowdhuri, \href{https://aclanthology.org/2022.wiesp-1.12}{Astro-m{T}5: Entity extraction from astrophysics literature using m{T}5 language model}, in: T.~Ghosal, S.~Blanco-Cuaresma, A.~Accomazzi, R.~M. Patton, F.~Grezes, T.~Allen (Eds.), Proceedings of the first Workshop on Information Extraction from Scientific Publications, Association for Computational Linguistics, Online, 2022, pp. 100--104.
\newline\urlprefix\url{https://aclanthology.org/2022.wiesp-1.12}

\bibitem{luan2018multi}
Y.~Luan, L.~He, M.~Ostendorf, H.~Hajishirzi, Multi-task identification of entities, relations, and coreference for scientific knowledge graph construction, arXiv preprint arXiv:1808.09602 (2018).

\bibitem{jain2020scirex}
S.~Jain, M.~van Zuylen, H.~Hajishirzi, I.~Beltagy, Scirex: A challenge dataset for document-level information extraction, arXiv preprint arXiv:2005.00512 (2020).

\bibitem{ghosh2023extracting}
M.~Ghosh, D.~Ganguly, P.~Basuchowdhuri, S.~K. Naskar, Extracting methodology components from ai research papers: A data-driven factored sequence labeling approach, in: Proceedings of the 32nd ACM International Conference on Information and Knowledge Management, 2023, pp. 3897--3901.

\bibitem{ghosh2023enhancing}
M.~Ghosh, D.~Ganguly, P.~Basuchowdhuri, S.~K. Naskar, Enhancing ai research paper analysis: Methodology component extraction using factored transformer-based sequence modeling approach, arXiv preprint arXiv:2311.03401 (2023).

\bibitem{tong2021multi}
Y.~Tong, Y.~Chen, X.~Shi, A multi-task approach for improving biomedical named entity recognition by incorporating multi-granularity information, in: Findings of the ACL: ACL-IJCNLP 2021, ACL, Online, 2021, pp. 4804--4813.

\bibitem{zhu2015aligning}
Y.~Zhu, R.~Kiros, R.~Zemel, R.~Salakhutdinov, R.~Urtasun, A.~Torralba, S.~Fidler, Aligning books and movies: Towards story-like visual explanations by watching movies and reading books, in: Proceedings of the IEEE international conference on computer vision, 2015, pp. 19--27.

\bibitem{akbik-etal-2018-contextual}
A.~Akbik, D.~Blythe, R.~Vollgraf, \href{https://aclanthology.org/C18-1139}{Contextual string embeddings for sequence labeling}, in: Proceedings of the 27th International Conference on Computational Linguistics, Association for Computational Linguistics, Santa Fe, New Mexico, USA, 2018, pp. 1638--1649.
\newline\urlprefix\url{https://aclanthology.org/C18-1139}

\bibitem{boudin2010combining}
F.~Boudin, J.-Y. Nie, J.~C. Bartlett, R.~Grad, P.~Pluye, M.~Dawes, Combining classifiers for robust pico element detection, BMC medical informatics and decision making 10~(1) (2010) 1--6.

\bibitem{huang2011classification}
K.-C. Huang, C.~C.-H. Liu, S.-S. Yang, F.~Xiao, J.-M. Wong, C.-C. Liao, I.-J. Chiang, Classification of pico elements by text features systematically extracted from pubmed abstracts, in: 2011 IEEE International Conference on Granular Computing, IEEE, 2011, pp. 279--283.

\bibitem{peters-etal-2018-deep}
M.~E. Peters, M.~Neumann, M.~Iyyer, M.~Gardner, C.~Clark, K.~Lee, L.~Zettlemoyer, Deep contextualized word representations, in: Proceedings of the 2018 Conference of the NAACL, ACL, New Orleans, Louisiana, 2018, pp. 2227--2237.

\bibitem{radford2018improving}
A.~Radford, K.~Narasimhan, T.~Salimans, I.~Sutskever, et~al., Improving language understanding by generative pre-training (2018).

\bibitem{devlin-etal-2019-bert}
J.~Devlin, M.-W. Chang, K.~Lee, K.~Toutanova, {BERT}: Pre-training of deep bidirectional transformers for language understanding, in: Proceedings of the 2019 ACL, Volume 1 (Long and Short Papers), ACL, Minneapolis, Minnesota, 2019, pp. 4171--4186.

\bibitem{lample2019cross}
G.~Lample, A.~Conneau, Cross-lingual language model pretraining, arXiv preprint arXiv:1901.07291 (2019).

\bibitem{yang2019xlnet}
Z.~Yang, Z.~Dai, Y.~Yang, J.~Carbonell, R.~R. Salakhutdinov, Q.~V. Le, Xlnet: Generalized autoregressive pretraining for language understanding, Advances in neural information processing systems 32 (2019).

\bibitem{sang2003introduction}
E.~F. Sang, F.~De~Meulder, Introduction to the conll-2003 shared task: Language-independent named entity recognition, arXiv preprint cs/0306050 (2003).

\bibitem{rajpurkar2016squad}
P.~Rajpurkar, J.~Zhang, K.~Lopyrev, P.~Liang, Squad: 100,000+ questions for machine comprehension of text, arXiv preprint arXiv:1606.05250 (2016).

\bibitem{liu-etal-2021-sent2span-span}
S.~Liu, Y.~Sun, B.~Li, W.~Wang, F.~T. Bourgeois, A.~G. Dunn, {S}ent2{S}pan: Span detection for {PICO} extraction in the biomedical text without span annotations, in: Findings of the ACL: EMNLP 2021, ACL, Punta Cana, Dominican Republic, 2021, pp. 1705--1715.

\bibitem{brockmeier2019improving}
A.~J. Brockmeier, M.~Ju, P.~Przyby{\l}a, S.~Ananiadou, Improving reference prioritisation with pico recognition, BMC medical informatics and decision making 19~(1) (2019) 1--14.

\bibitem{zhang2020unlocking}
T.~Zhang, Y.~Yu, J.~Mei, Z.~Tang, X.~Zhang, S.~Li, Unlocking the power of deep pico extraction: Step-wise medical ner identification, arXiv preprint arXiv:2005.06601 (2020).

\bibitem{ghosh2024blinktextsubscriptlstm}
M.~Ghosh, S.~Mukherjee, P.~Santra, G.~Na, P.~Basuchowdhuri, Blinktextsubscriptlstm: Biolinkbert and lstm based approach for extraction of pico frame from clinical trial text, in: Proceedings of the 7th Joint International Conference on Data Science \& Management of Data (11th ACM IKDD CODS and 29th COMAD), 2024, pp. 227--231.

\bibitem{dhrangadhariya-muller-2022-distant}
A.~Dhrangadhariya, H.~M{\"u}ller, {DISTANT}-{CTO}: A zero cost, distantly supervised approach to improve low-resource entity extraction using clinical trials literature, in: Proceedings of the 21st Workshop on Biomedical Language Processing, ACL, Dublin, Ireland, 2022, pp. 345--358.

\bibitem{giannakopoulos-etal-2017-unsupervised}
A.~Giannakopoulos, C.~Musat, A.~Hossmann, M.~Baeriswyl, Unsupervised aspect term extraction with {B}-{LSTM} {\&} {CRF} using automatically labelled datasets, in: Proceedings of the 8th Workshop on Computational Approaches to Subjectivity, Sentiment and Social Media Analysis, ACL, Copenhagen, Denmark, 2017, pp. 180--188.

\bibitem{mullenbach-etal-2018-explainable}
J.~Mullenbach, S.~Wiegreffe, J.~Duke, J.~Sun, J.~Eisenstein, Explainable prediction of medical codes from clinical text, in: Proceedings of the 2018 Conference of NAACL, ACL, New Orleans, Louisiana, 2018, pp. 1101--1111.

\bibitem{brown2020language}
T.~Brown, B.~Mann, N.~Ryder, M.~Subbiah, J.~D. Kaplan, P.~Dhariwal, A.~Neelakantan, P.~Shyam, G.~Sastry, A.~Askell, et~al., Language models are few-shot learners, Advances in neural information processing systems 33 (2020) 1877--1901.

\bibitem{rae2021scaling}
J.~W. Rae, S.~Borgeaud, T.~Cai, K.~Millican, J.~Hoffmann, F.~Song, J.~Aslanides, S.~Henderson, R.~Ring, S.~Young, et~al., Scaling language models: Methods, analysis \& insights from training gopher, arXiv preprint arXiv:2112.11446 (2021).

\bibitem{smith2022using}
S.~Smith, M.~Patwary, B.~Norick, P.~LeGresley, S.~Rajbhandari, J.~Casper, Z.~Liu, S.~Prabhumoye, G.~Zerveas, V.~Korthikanti, et~al., Using deepspeed and megatron to train megatron-turing nlg 530b, a large-scale generative language model, arXiv preprint arXiv:2201.11990 (2022).

\bibitem{hoffmann2022training}
J.~Hoffmann, S.~Borgeaud, A.~Mensch, E.~Buchatskaya, T.~Cai, E.~Rutherford, D.~d.~L. Casas, L.~A. Hendricks, J.~Welbl, A.~Clark, et~al., Training compute-optimal large language models, arXiv preprint arXiv:2203.15556 (2022).

\bibitem{chowdhery2022palm}
A.~Chowdhery, S.~Narang, J.~Devlin, M.~Bosma, G.~Mishra, A.~Roberts, P.~Barham, H.~W. Chung, C.~Sutton, S.~Gehrmann, et~al., Palm: Scaling language modeling with pathways, arXiv preprint arXiv:2204.02311 (2022).

\bibitem{hegselmann2022tabllm}
S.~Hegselmann, A.~Buendia, H.~Lang, M.~Agrawal, X.~Jiang, D.~Sontag, Tabllm: Few-shot classification of tabular data with large language models, arXiv preprint arXiv:2210.10723 (2022).

\bibitem{vilar2022prompting}
D.~Vilar, M.~Freitag, C.~Cherry, J.~Luo, V.~Ratnakar, G.~Foster, Prompting palm for translation: Assessing strategies and performance, arXiv preprint arXiv:2211.09102 (2022).

\bibitem{perez2021true}
E.~Perez, D.~Kiela, K.~Cho, True few-shot learning with language models, Advances in neural information processing systems 34 (2021) 11054--11070.

\bibitem{pietrzak2021story}
B.~Pietrzak, B.~Swanson, K.~Mathewson, M.~Dinculescu, S.~Chen, Story centaur: Large language model few shot learning as a creative writing tool (2021).

\bibitem{wei2021finetuned}
J.~Wei, M.~Bosma, V.~Y. Zhao, K.~Guu, A.~W. Yu, B.~Lester, N.~Du, A.~M. Dai, Q.~V. Le, Finetuned language models are zero-shot learners, arXiv preprint arXiv:2109.01652 (2021).

\bibitem{raffel2020exploring}
C.~Raffel, N.~Shazeer, A.~Roberts, K.~Lee, S.~Narang, M.~Matena, Y.~Zhou, W.~Li, P.~J. Liu, Exploring the limits of transfer learning with a unified text-to-text transformer, The Journal of Machine Learning Research 21~(1) (2020) 5485--5551.

\bibitem{gururangan2018annotation}
S.~Gururangan, S.~Swayamdipta, O.~Levy, R.~Schwartz, S.~R. Bowman, N.~A. Smith, Annotation artifacts in natural language inference data, arXiv preprint arXiv:1803.02324 (2018).

\bibitem{roberts2020much}
A.~Roberts, C.~Raffel, N.~Shazeer, How much knowledge can you pack into the parameters of a language model?, arXiv preprint arXiv:2002.08910 (2020).

\bibitem{guu2020retrieval}
K.~Guu, K.~Lee, Z.~Tung, P.~Pasupat, M.~Chang, Retrieval augmented language model pre-training (2020) 3929--3938.

\bibitem{radford2019language}
A.~Radford, J.~Wu, R.~Child, D.~Luan, D.~Amodei, I.~Sutskever, et~al., Language models are unsupervised multitask learners, OpenAI blog 1~(8) (2019) 9.

\bibitem{lu2021fantastically}
Y.~Lu, M.~Bartolo, A.~Moore, S.~Riedel, P.~Stenetorp, Fantastically ordered prompts and where to find them: Overcoming few-shot prompt order sensitivity, arXiv preprint arXiv:2104.08786 (2021).

\bibitem{rubin2021learning}
O.~Rubin, J.~Herzig, J.~Berant, Learning to retrieve prompts for in-context learning, arXiv preprint arXiv:2112.08633 (2021).

\bibitem{wang2023gpt}
S.~Wang, X.~Sun, X.~Li, R.~Ouyang, F.~Wu, T.~Zhang, J.~Li, G.~Wang, Gpt-ner: Named entity recognition via large language models, arXiv preprint arXiv:2304.10428 (2023).

\bibitem{zhang2023promptner}
M.~Zhang, H.~Yan, Y.~Zhou, X.~Qiu, Promptner: A prompting method for few-shot named entity recognition via k nearest neighbor search, arXiv preprint arXiv:2305.12217 (2023).

\bibitem{pakhale2023comprehensive}
K.~Pakhale, Comprehensive overview of named entity recognition: Models, domain-specific applications and challenges, arXiv preprint arXiv:2309.14084 (2023).

\bibitem{ashok2023promptner}
D.~Ashok, Z.~C. Lipton, Promptner: Prompting for named entity recognition, arXiv preprint arXiv:2305.15444 (2023).

\bibitem{chung2022scaling}
H.~W. Chung, L.~Hou, S.~Longpre, B.~Zoph, Y.~Tay, W.~Fedus, E.~Li, X.~Wang, M.~Dehghani, S.~Brahma, et~al., Scaling instruction-finetuned language models, arXiv preprint arXiv:2210.11416 (2022).

\bibitem{sanh2021multitask}
V.~Sanh, A.~Webson, C.~Raffel, S.~H. Bach, L.~Sutawika, Z.~Alyafeai, A.~Chaffin, A.~Stiegler, T.~L. Scao, A.~Raja, M.~Dey, M.~S. Bari, C.~Xu, U.~Thakker, S.~S. Sharma, E.~Szczechla, T.~Kim, G.~Chhablani, N.~Nayak, D.~Datta, J.~Chang, M.~T.-J. Jiang, H.~Wang, M.~Manica, S.~Shen, Z.~X. Yong, H.~Pandey, R.~Bawden, T.~Wang, T.~Neeraj, J.~Rozen, A.~Sharma, A.~Santilli, T.~Fevry, J.~A. Fries, R.~Teehan, S.~Biderman, L.~Gao, T.~Bers, T.~Wolf, A.~M. Rush, Multitask prompted training enables zero-shot task generalization (2021).
\newblock \href {http://arxiv.org/abs/2110.08207} {\path{arXiv:2110.08207}}.

\bibitem{wang2022super}
Y.~Wang, S.~Mishra, P.~Alipoormolabashi, Y.~Kordi, A.~Mirzaei, A.~Arunkumar, A.~Ashok, A.~S. Dhanasekaran, A.~Naik, D.~Stap, et~al., Super-naturalinstructions: Generalization via declarative instructions on 1600+ nlp tasks, arXiv preprint arXiv:2204.07705 (2022).

\bibitem{ouyang2022training}
L.~Ouyang, J.~Wu, X.~Jiang, D.~Almeida, C.~L. Wainwright, P.~Mishkin, C.~Zhang, S.~Agarwal, K.~Slama, A.~Ray, J.~Schulman, J.~Hilton, F.~Kelton, L.~Miller, M.~Simens, A.~Askell, P.~Welinder, P.~Christiano, J.~Leike, R.~Lowe, Training language models to follow instructions with human feedback (2022).
\newblock \href {http://arxiv.org/abs/2203.02155} {\path{arXiv:2203.02155}}.

\bibitem{alpaca}
R.~Taori, I.~Gulrajani, T.~Zhang, Y.~Dubois, X.~Li, C.~Guestrin, P.~Liang, T.~B. Hashimoto, Stanford alpaca: An instruction-following llama model, \url{https://github.com/tatsu-lab/stanford_alpaca} (2023).

\bibitem{vicuna-2023}
W.-L. Chiang, Z.~Li, Z.~Lin, Y.~Sheng, Z.~Wu, H.~Zhang, L.~Zheng, S.~Zhuang, Y.~Zhuang, J.~E. Gonzalez, I.~Stoica, E.~P. Xing, \href{https://vicuna.lmsys.org}{Vicuna: An open-source chatbot impressing gpt-4 with 90\%* chatgpt quality} (March 2023).
\newline\urlprefix\url{https://vicuna.lmsys.org}

\bibitem{peng2023instruction}
B.~Peng, C.~Li, P.~He, M.~Galley, J.~Gao, Instruction tuning with gpt-4, arXiv preprint arXiv:2304.03277 (2023).

\bibitem{wang2023far}
Y.~Wang, H.~Ivison, P.~Dasigi, J.~Hessel, T.~Khot, K.~R. Chandu, D.~Wadden, K.~MacMillan, N.~A. Smith, I.~Beltagy, H.~Hajishirzi, How far can camels go? exploring the state of instruction tuning on open resources (2023).
\newblock \href {http://arxiv.org/abs/2306.04751} {\path{arXiv:2306.04751}}.

\bibitem{gudibande2023false}
A.~Gudibande, E.~Wallace, C.~Snell, X.~Geng, H.~Liu, P.~Abbeel, S.~Levine, D.~Song, The false promise of imitating proprietary llms (2023).
\newblock \href {http://arxiv.org/abs/2305.15717} {\path{arXiv:2305.15717}}.

\bibitem{wang2023instructuie}
X.~Wang, W.~Zhou, C.~Zu, H.~Xia, T.~Chen, Y.~Zhang, R.~Zheng, J.~Ye, Q.~Zhang, T.~Gui, J.~Kang, J.~Yang, S.~Li, C.~Du, Instructuie: Multi-task instruction tuning for unified information extraction (2023).
\newblock \href {http://arxiv.org/abs/2304.08085} {\path{arXiv:2304.08085}}.

\bibitem{zhou2023universalner}
W.~Zhou, S.~Zhang, Y.~Gu, M.~Chen, H.~Poon, Universalner: Targeted distillation from large language models for open named entity recognition, arXiv preprint arXiv:2308.03279 (2023).

\bibitem{min-etal-2022-metaicl}
S.~Min, M.~Lewis, L.~Zettlemoyer, H.~Hajishirzi, \href{https://aclanthology.org/2022.naacl-main.201}{{M}eta{ICL}: Learning to learn in context}, in: Proceedings of the 2022 Conference of the North American Chapter of the Association for Computational Linguistics: Human Language Technologies, Association for Computational Linguistics, Seattle, United States, 2022, pp. 2791--2809.
\newblock \href {https://doi.org/10.18653/v1/2022.naacl-main.201} {\path{doi:10.18653/v1/2022.naacl-main.201}}.
\newline\urlprefix\url{https://aclanthology.org/2022.naacl-main.201}

\bibitem{zhang2023alpacareinstructiontuned}
X.~Zhang, C.~Tian, X.~Yang, L.~Chen, Z.~Li, L.~R. Petzold, Alpacare:instruction-tuned large language models for medical application (2023).
\newblock \href {http://arxiv.org/abs/2310.14558} {\path{arXiv:2310.14558}}.

\bibitem{wu-etal-2023-openicl}
Z.~Wu, Y.~Wang, J.~Ye, Z.~Wu, J.~Feng, J.~Xu, Y.~Qiao, \href{https://aclanthology.org/2023.acl-demo.47}{{O}pen{ICL}: An open-source framework for in-context learning}, in: D.~Bollegala, R.~Huang, A.~Ritter (Eds.), Proceedings of the 61st Annual Meeting of the Association for Computational Linguistics (Volume 3: System Demonstrations), Association for Computational Linguistics, Toronto, Canada, 2023, pp. 489--498.
\newblock \href {https://doi.org/10.18653/v1/2023.acl-demo.47} {\path{doi:10.18653/v1/2023.acl-demo.47}}.
\newline\urlprefix\url{https://aclanthology.org/2023.acl-demo.47}

\bibitem{fu2022effectiveness}
Z.~Fu, H.~Yang, A.~M.-C. So, W.~Lam, L.~Bing, N.~Collier, On the effectiveness of parameter-efficient fine-tuning (2022).
\newblock \href {http://arxiv.org/abs/2211.15583} {\path{arXiv:2211.15583}}.

\bibitem{lora}
E.~J. Hu, Y.~Shen, P.~Wallis, Z.~Allen-Zhu, Y.~Li, S.~Wang, W.~Chen, Lora: Low-rank adaptation of large language models, ArXiv abs/2106.09685 (2021).

\bibitem{krona}
A.~Edalati, M.~S. Tahaei, I.~Kobyzev, V.~Nia, J.~J. Clark, M.~Rezagholizadeh, Krona: Parameter efficient tuning with kronecker adapter, ArXiv abs/2212.10650 (2022).

\bibitem{fang2023mol}
Y.~Fang, X.~Liang, N.~Zhang, K.~Liu, R.~Huang, Z.~Chen, X.~Fan, H.~Chen, Mol-instructions: A large-scale biomolecular instruction dataset for large language models, arXiv preprint arXiv:2306.08018 (2023).

\bibitem{loshchilov2018decoupled}
I.~Loshchilov, F.~Hutter, Decoupled weight decay regularization, in: International Conference on Learning Representations, 2019.

\bibitem{conneau-etal-2020-unsupervised}
A.~Conneau, K.~Khandelwal, N.~Goyal, V.~Chaudhary, G.~Wenzek, F.~Guzm{\'a}n, E.~Grave, M.~Ott, L.~Zettlemoyer, V.~Stoyanov, Unsupervised cross-lingual representation learning at scale, in: Proceedings of the 58th Annual Meeting of the ACL, ACL, Online, 2020, pp. 8440--8451.

\bibitem{yang-etal-2018-sgm}
P.~Yang, X.~Sun, W.~Li, S.~Ma, W.~Wu, H.~Wang, \href{https://aclanthology.org/C18-1330}{{SGM}: Sequence generation model for multi-label classification}, in: E.~M. Bender, L.~Derczynski, P.~Isabelle (Eds.), Proceedings of the 27th International Conference on Computational Linguistics, Association for Computational Linguistics, Santa Fe, New Mexico, USA, 2018, pp. 3915--3926.
\newline\urlprefix\url{https://aclanthology.org/C18-1330}

\bibitem{lee2020biobert}
J.~Lee, W.~Yoon, S.~Kim, D.~Kim, S.~Kim, C.~H. So, J.~Kang, Biobert: a pre-trained biomedical language representation model for biomedical text mining, Bioinformatics 36~(4) (2020) 1234--1240.

\bibitem{touvron2023llama}
H.~Touvron, T.~Lavril, G.~Izacard, X.~Martinet, M.-A. Lachaux, T.~Lacroix, B.~Rozi{\`e}re, N.~Goyal, E.~Hambro, F.~Azhar, et~al., Llama: Open and efficient foundation language models, arXiv preprint arXiv:2302.13971 (2023).

\bibitem{malkov2018efficient}
Y.~A. Malkov, D.~A. Yashunin, Efficient and robust approximate nearest neighbor search using hierarchical navigable small world graphs (2018).
\newblock \href {http://arxiv.org/abs/1603.09320} {\path{arXiv:1603.09320}}.

\bibitem{wuself}
Z.~Wu, Y.~Wang, J.~Ye, L.~Kong, Self-adaptive in-context learning: An information compression perspective for in-context example selection and ordering.

\end{thebibliography}

% \end{thebibliography}
\end{document}